\documentclass[11pt]{article}
\usepackage[a4paper,margin=1in]{geometry}

\usepackage[numbers,square]{natbib}

\usepackage[T1]{fontenc}
\usepackage{graphicx}
\usepackage{amsmath}
\usepackage{newtxtext}   
\usepackage{newtxmath}   
\usepackage{booktabs}
\usepackage{siunitx}
\usepackage{xspace}
\usepackage{etoolbox}   
\usepackage{array}
\usepackage{makecell}
\usepackage{microtype}
\usepackage{subcaption}          
\usepackage[section]{placeins}   

\newcommand{\tblwidth}{\linewidth}
\newcolumntype{L}{@{\extracolsep{\fill}}l}
\newcolumntype{R}{@{\extracolsep{\fill}}r}
\newcolumntype{C}{@{\extracolsep{\fill}}c}

\def\tsc#1{\csdef{#1}{\textsc{\lowercase{#1}}\xspace}}
\tsc{RLHF} \tsc{NeRF} \tsc{GAN} \tsc{EG3D} \tsc{FFHQ}

\usepackage[colorlinks=true,linkcolor=blue,citecolor=blue,urlcolor=blue]{hyperref}

\begin{document}
\title{Sculpting NeRF Geometry: Human-Preference Fine-Tuning of a 3D-Aware Face GAN}
\author{Archer~Moore\thanks{Corresponding author: \texttt{archerplmoore@gmail.com}} \quad Mingming~Gong \quad Liam~Hodgkinson \\
  School of Mathematics and Statistics, \\
  The University of Melbourne, Parkville, VIC 3010, Australia}
\date{}
\maketitle

\begin{abstract}
Reinforcement learning from human feedback (RLHF) for 3D generation is now established across a number of works, but most existing pipelines optimise explicit surface representations, often by converting radiance fields into meshes and training heavily on surface-supervised data. We instead fine-tune a pretrained 3D-aware generative model directly from a learned reward over radiance-field density ($\sigma$) values, with no externally supplied mesh or shape prior. The reward model requires no pretraining, trains easily on a small set of preference samples, and yields robust improvement in 3D geometry. Working on an unconditional 3D-aware face GAN (EG3D), our reward reads the continuous 3D density field of the neural radiance field (NeRF) directly and supplies a geometry-only learning signal, requiring neither text conditioning, mesh extraction, nor multi-view rendering. A density-consistency constraint keeps the 2D appearance qualitatively similar while the geometry is reshaped, at a measurable but bounded distributional cost (FID-50k rises from $4.09$ to $6.66$): the fine-tuned generator, trained from the preferences of a single annotator as a proof of concept, produces face geometries preferred by users in $74.4\%$ of pairwise comparisons.
\end{abstract}


\section{Introduction}\label{sec:intro}

\begin{figure}[htbp]
  \centering
  \includegraphics[width=0.98\linewidth]{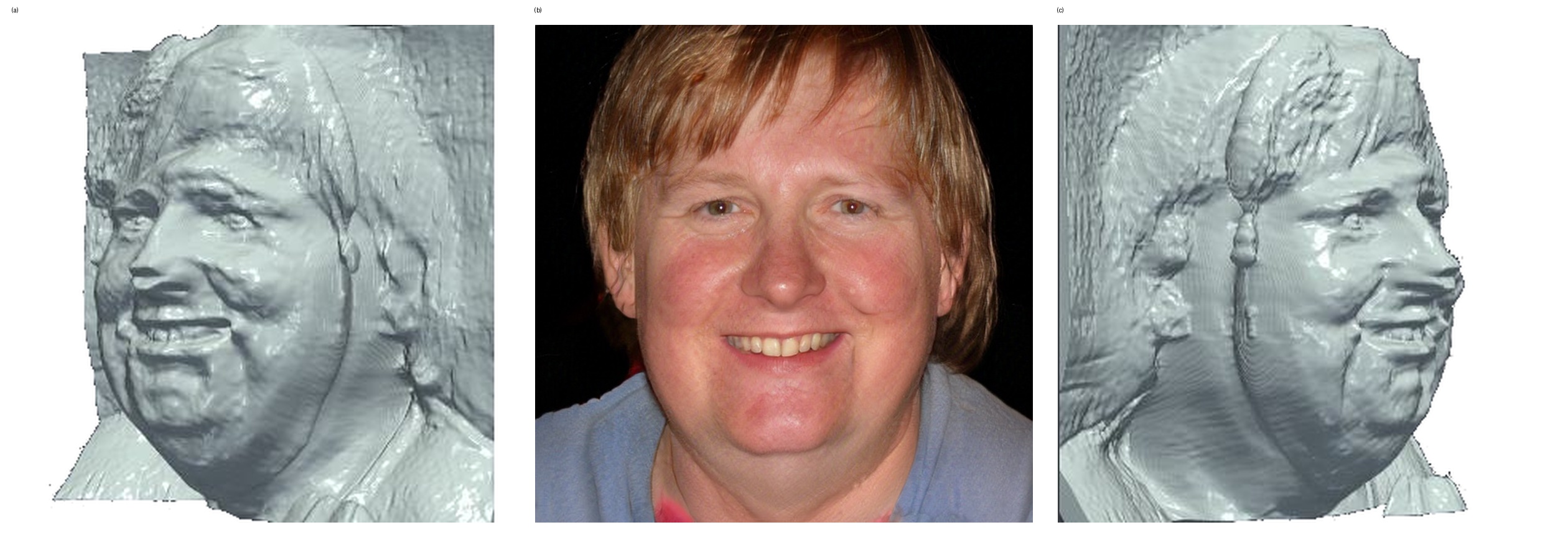}
  \caption{Appearance and geometry for a fixed latent code sampled from EG3D 3D-aware face generator. The rendered image appears plausible, but the underlying mesh exhibits unrealistic grooves, bumps and side-face artefacts.}
  \label{fig:appearance_geometry_gap}
\end{figure}

Generative computer vision models trained on 2D image collections have advanced considerably in recent years, achieving photorealistic quality in novel-view synthesis \citep{chan2020pigan,chan2022eg3d,karras2019stylegan,karras2021aliasfree}. Extending these works, 3D-aware models employ an image-rendering process to infer shape and appearance via 2D image reconstruction from 3D features parameterised by a neural network. This allows novel images to be rendered with independent control of the camera viewpoint and the underlying 3D geometry to be extracted as a by-product. Such methods have captured wide research interest because they derive 3D information from unstructured 2D image collections, but improving 3D shape quality remains an open problem: despite realistic image outputs, the recovered 3D shapes often contain unrealistic discontinuities or irregular geometries. This is particularly visible on models trained on single-category image collections of human faces such as Flickr-Faces-HQ (FFHQ) \citep{karras2019ffhq}. Even with a state-of-the-art 3D GAN such as EG3D \citep{chan2022eg3d}, geometric defects are routinely observable on the nose and around the sides of the face. We explore an approach for fine-tuning the geometry based on human preferences of 3D shape quality alone, without using any further information such as a 3D mesh prior \citep{gerig2018morphable}. Figure~\ref{fig:appearance_geometry_gap} makes the core problem concrete: realistic 2D appearance does not guarantee realistic recovered 3D geometry.

\begin{figure}[htbp]
  \centering
  \includegraphics[width=0.80\linewidth]{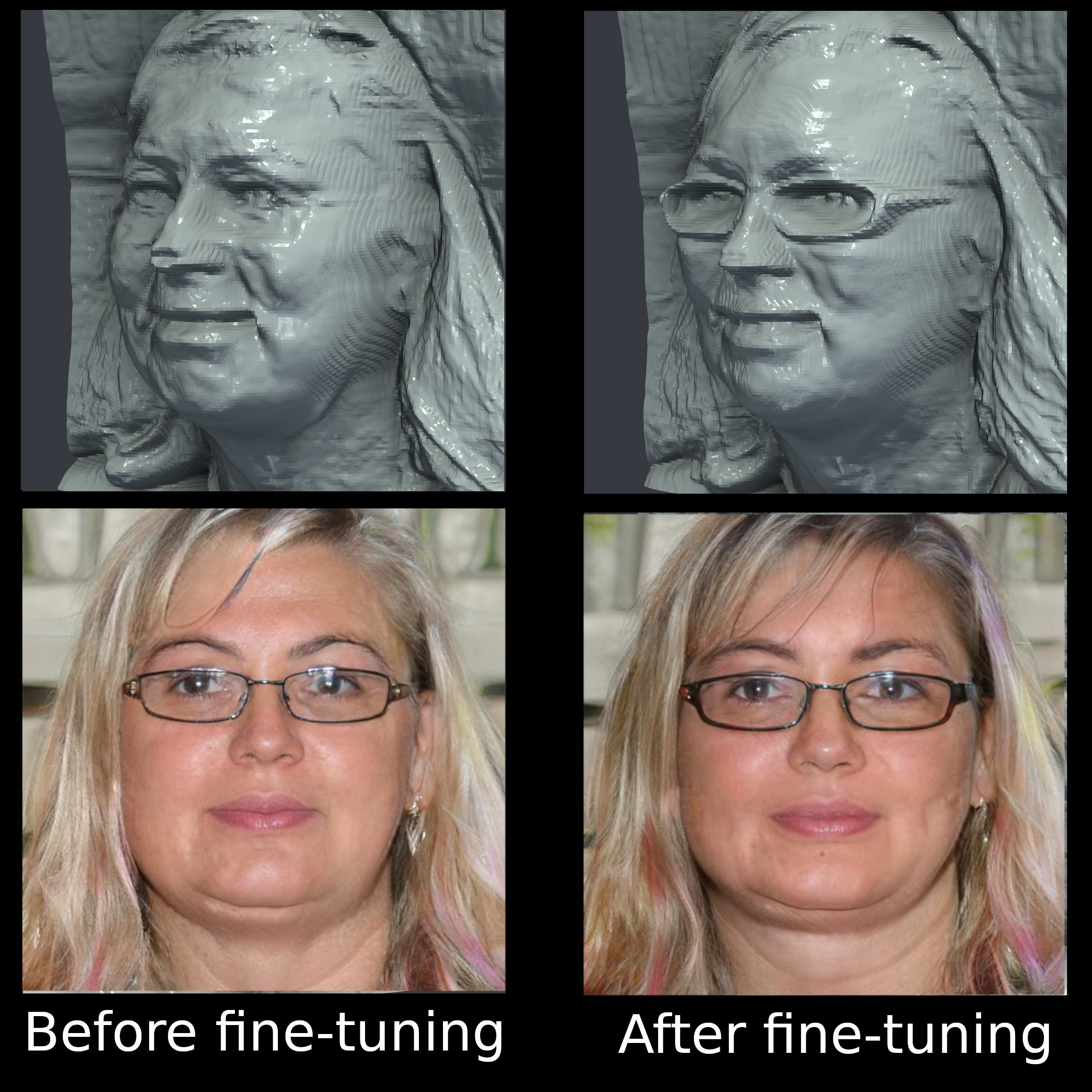}
  \caption{Geometry and appearance before (left) and after (right) fine-tuning with human feedback, for a representative seed. Top: the $\sigma$-level-$10$ marching-cubes mesh; bottom: the RGB render. Before fine-tuning the glasses are present in the RGB render but absent from the extracted geometry; after fine-tuning they appear in the geometry as well. Identity and overall appearance are preserved between the two RGB renders, with minor differences discernible -- slightly darker lighting, a few more hair strands across the forehead, marginally stronger purple highlights, and thicker glasses.}
  \label{fig:before_after}
\end{figure}

A wave of preference-driven 3D generative methods has emerged since the original framing of this question \citep{ye2024dreamreward,zhou2025dreamdpo,zhao2025deepmesh,liu2025meshrft,liu2025dreamreward,zou2026dreamcs,wang2025mvreward,liu2025dreamalign,huang2024humannorm}. With few exceptions, these methods condition on a natural-language prompt and either score multi-view rendered images or operate on mesh tokens. Our setting is structurally different. We operate on an unconditional 3D-aware face GAN, our reward model scores the NeRF density volume of the generator directly - without rendering or mesh extraction - and a density-consistency constraint keeps the 2D appearance qualitatively similar during fine-tuning, at a small but measurable distributional cost. This matters concretely for preference tuning: \citet{chen2026beyondprompts} show on a recent text-to-3D backbone that there are regions of latent space (``sink traps'') where editing the prompt no longer changes the produced geometry, so a text-conditioned reward can be left steering a signal the generator has stopped responding to; by contrast, the same backbone's unconditional prior remains useful for inversion and editing in those regimes. Operating without a prompt, as we do, sidesteps this failure mode. The method is inspired by InstructGPT-style preference optimisation \citep{ouyang2022instructgpt}, but rather than improving a conditional estimate $r_{\theta}(x,y)$ over a prompt $x$ and response $y$, we learn an unconditional critic $r_{\theta}(x_{3D})$ that evaluates the quality of 3D features $x_{3D}$ extracted from the pretrained generator and use it as a fine-tuning signal. Figure~\ref{fig:before_after} shows representative geometries before and after fine-tuning: face sides and the nose region are smoothed and made more plausible while the front-facing geometry and 2D appearance remain qualitatively similar. The main contributions of this work are as follows:
\begin{enumerate}
\item We show that it is possible to learn 3D shape quality from direct human rankings of a small number of generator outputs. This simplifies existing works requiring refined assessments of multiple shape regions or Likert-scale ratings in multiple dimensions \citep{liu2022pcqa,zhang2025mate3d,ye2024dreamreward} to derive quality scores.
\item A quality-scoring module $r_{\theta}$ operating directly on the 3D density representation is developed. It requires neither language-prompt conditioning \citep{ye2024dreamreward,zou2026dreamcs} nor colour information \citep{ye2024dreamreward,wang2025mvreward}, and it does not rely on pretraining over mesh collections or on explicit surface constraints.
\item According to user studies, 3D shapes extracted from EG3D after fine-tuning with human output are preferred over their original outputs in $74{.}4\%$ of pairwise comparisons.
\end{enumerate}

The remainder of this paper is organised as follows. Section~\ref{sec:related} reviews 3D-aware generative models, the fine-tuning of generative models with human feedback, and 3D shape quality assessment, and positions our contribution against the recent wave of 3D RLHF methods. Section~\ref{sec:method} develops the reward-model architecture and the fine-tuning procedure. Section~\ref{sec:results} reports reward-model ablations, fine-tuning results on EG3D, an external user study, and an analysis of intermediate representations via SHAP \citep{lundberg2017shap}. Sections~\ref{sec:discussion}~and~\ref{sec:conclusion} discuss limitations and implications. Code, the trained $\sigma_{XYZ}$ reward model, and the fine-tuned EG3D checkpoint are available at \url{https://github.com/apmoore499/eg3d-rlhf-geometry}.

\section{Related Work}\label{sec:related}

\subsection{3D-aware generative models from 2D images}\label{sec:rel:3dgen}

Generative models that infer 3D representations from 2D image collections have matured rapidly since neural radiance fields (NeRFs) were introduced as a volumetric scene representation \citep{mildenhall2020nerf}. Early 3D-aware generative adversarial networks (GANs) embedded a NeRF or signed-distance field inside the generator and rendered images via volume rendering \citep{schwarz2020graf, chan2020pigan,gu2021stylenerf}. EG3D introduced a triplane representation that decoupled feature storage from rendering cost, achieving state-of-the-art FID on FFHQ \citep{chan2022eg3d,karras2019ffhq}, learning an implicit density volume which our method exploits. A series of follow-up works extends EG3D-style face generators to wider viewing distributions: PanoHead \citep{an2023panohead} introduces a tri-grid representation for $360^{\circ}$ heads; SphereHead \citep{li2024spherehead} replaces the axis-aligned triplane with a spherical-plane parameterisation; and HyPlaneHead \citep{li2026hyplanehead} further consolidates spherical-plane features into a single fused dimension. Geometry-aware regularisers \citep{shi2022geod} and pose-distribution augmentation \citep{skorokhodov2022epigraf} have been used to combat the front-back entanglement and concave-nose artefacts that motivate our reward-based approach. We refer the reader to \citet{shi2022survey3dgen} for a broader survey.

Recent image-to-3D pipelines bypass volumetric rendering and predict explicit meshes or compact latents. CraftsMan3D \citep{li2025craftsman3d} couples a 3D-native diffusion prior with an interactive geometry refiner. Hi3DGen \citep{ye2025hi3dgen} bridges images to high-fidelity 3D geometry through predicted normal maps. Trellis \citep{xiang2024trellis} and Trellis~2 \citep{xiang2025trellis2} introduce structured 3D latents that compress sparse features with arbitrary topology with an efficient voxel-grid encoding, enabling text- and image-conditional 3D generation at high resolution. Variants such as GaussianCube \citep{zhang2024gaussiancube} and LN3Diff \citep{lan2024ln3diff} explore latent diffusion in alternative 3D representations.

NeRF approaches are popular for human face models. Recent works in this area include Gen3D-Face for generalisable single-image 3D face generation via multi-view diffusion with input-conditioned mesh estimation \citet{wang2026singleimageanyface}, unified NeRF-mesh joint optimisation \citep{neupane2026nerfmesh}, using audio information to drive facial portrait generation \citep{yang2026facialtalk}, and silhouette-initialised radiance fields from sparse inputs \citep{lai2025fastradiance}. Our contribution differs from this line in that we do not propose a new 3D representation or reconstruction pipeline, but instead a reward-based fine-tuning procedure that acts on an existing pretrained 3D-aware GAN backbone.

\subsection{Fine-tuning generative models with human feedback}\label{sec:rel:rlhf}

Reinforcement learning from human feedback (RLHF) was originally formalised for Markov decision processes by \citet{christiano2017drlhf} and brought to generative sequence modelling through the summarisation-from-feedback line \citep{stiennon2020summarise} and InstructGPT \citep{ouyang2022instructgpt}, where a learned pairwise reward model is optimised against the generator with Proximal Policy Optimisation \citep{schulman2017ppo}. Recent surveys give a broader overview of reinforcement learning for visual and 3D generation \citep{liang2026rlvisgen,wu2025rlsurvey}.

Adaptation of these ideas to 2D image generation began with preference-based image generation \citep{kazemi2020pbig} and now includes ImageReward \citep{xu2024imagereward}, with multiple more recent works motivating the use of human feedback of 2D appearance as a signal to guide inversion, editing, or model training \citep{kirstain2023pickapic,helbling2023prefgen,tang2023zerothorder,huang2024hutumotion}.

Extension of preference-based fine-tuning to 3D generative models is considerably more recent. DreamReward, a 3D reward model trained on 25\,000 expert pairwise comparisons of multi-view renderings, demonstrates refinement of text-to-3D pipelines via score-distillation sampling \citep{ye2024dreamreward}. DreamControl \citep{huang2024dreamcontrol} and HumanNorm \citep{huang2024humannorm} address related text-to-3D control problems through self-priors and normal-aware diffusion respectively. The DPO formulation was carried into 3D by DreamDPO \citep{zhou2025dreamdpo}, which operates on pairwise multi-view comparisons; into autoregressive mesh generation by DeepMesh \citep{zhao2025deepmesh}, which mixes human preferences with topological metrics over 5\,000 preference pairs; and into fine-grained mesh post-training by Mesh-RFT \citep{liu2025meshrft}, which applies masked DPO at the face level. \citet{liu2025dreamreward} subsequently extended the DreamReward framework to image-to-3D and 4D settings. Most recently, DreamCS \citep{zou2026dreamcs} trains a 3D reward model in mesh feature space using a Cauchy--Schwarz divergence that admits unpaired preference data.

\citet{wang2025mvreward} report a multi-view reward model trained on 16\,000 expert comparisons that aligns multi-view diffusion models with human preferences, while DreamAlign \citep{liu2025dreamalign} dispenses with an explicit reward model and instead injects preferences through LoRA-augmented text prompts. Nabla-R2D3 \citep{liu2025nablar2d3} aligns 3D-native diffusion models using 2D rewards through a GFlowNet-style score-matching objective; in contrast, our reward reads the generator's density field directly rather than rendered 2D imagery.

Our setting departs from these works in three connected respects. They condition on a natural-language prompt, whereas our generator is unconditional. Their reward models score either rendered 2D imagery (DreamReward, MVReward, DreamDPO) or explicit mesh tokens (DreamCS, DeepMesh, Mesh-RFT), whereas ours reads the implicit density field $\sigma_{XYZ}$ of a NeRF directly. And because their reward is conditioned on a text prompt, it reshapes geometry jointly with prompt-conditioned appearance, whereas our reward is prompt-free and a density-consistency constraint keeps the 2D appearance qualitatively similar, at bounded cost, while the geometry alone is selectively improved. We adopt the InstructGPT pairwise reward formulation as the basis of our loss, replacing PPO with a modified GAN-loop update for compatibility with the pretrained EG3D backbone.

\subsection{3D shape quality assessment}\label{sec:rel:quality}

Independent of the generative-modelling literature, a body of work addresses no-reference 3D quality assessment as a perceptual prediction task. Recent text-to-3D evaluation benchmarks \citep{zhang2025mate3d,wu2024gpt} score multi-dimensional perceptual quality across text-prompt categories, either by collecting large-scale subjective annotations or by using a vision-language model as a judge. Closer to a learned quality signal, large pretrained mesh-language models are increasingly used to filter 3D assets by quality: DreamCS, for instance, scores meshes curated from Cap3D with LLaMA-Mesh \citep{wang2024llamamesh} on geometric fidelity, semantic alignment and structural plausibility, refined by human verification, to label preferred versus dispreferred examples for its reward model \citep{zou2026dreamcs}. Such methods are not designed to act directly on a generator's parameters, but several of their backbone choices transfer naturally to our reward-model architecture sweep. Our reward model extends the no-reference paradigm in two ways: its inputs are derived from a generator's internal density field rather than a sampled or scanned mesh, removing the dependence on a discretisation step; and its outputs are differentiable with respect to the generator's parameters, permitting end-to-end fine-tuning.

\subsection{Positioning of our contribution}\label{sec:rel:position}

Several features position our contribution within the landscape reviewed above. Our reward model $r_{\theta}\!: \sigma_{XYZ} \to s \in \mathbb{R}$ operates directly on the NeRF density volume of the generator, in contrast to other contemporary 3D preference-tuning methods, which evaluate either rendered multi-view images \citep{ye2024dreamreward,zhou2025dreamdpo,wang2025mvreward} or explicit mesh tokens \citep{zou2026dreamcs,zhao2025deepmesh,liu2025meshrft}. The pipeline is also unconditional: no text prompt enters the reward model or the fine-tuning loop, whereas every preference-driven 3D method published since \citet{ye2024dreamreward} conditions on a natural-language description that reshapes the reward signal jointly with appearance. The training corpus is correspondingly modest - $4{,}346$ pairwise comparisons from a single annotator, comparable to the $5{,}000$ paired samples of DeepMesh \citep{zhao2025deepmesh} and several times smaller than the $16{,}000$ expert comparisons of MVReward \citep{wang2025mvreward} and $25{,}000$ of DreamReward \citep{ye2024dreamreward}. The optimisation is a modified GAN-loop update over the original EG3D parameters, in contrast to score-distillation sampling \citep{ye2024dreamreward,huang2024humannorm}, optimisation-time guidance \citep{liu2025dreamalign}, and DPO objectives over discrete tokens \citep{zhou2025dreamdpo,liu2025meshrft,zhao2025deepmesh}. The Cauchy--Schwarz divergence machinery introduced by \citet{zou2026dreamcs} addresses the difficulty of comparing preference pairs across distinct text prompts; since our setting is unconditional and the preference data lie within a single prompt-free generator distribution, paired examples are trivially obtainable and the standard pairwise loss in Equation~\eqref{eq:lw} suffices.

DreamCS \citep{zou2026dreamcs} is in some respects a more general result, being shape-agnostic and showing evidence of cross-backbone transfer; methods in this domain often combine several stages of feedback, building on large pretrained mesh--language models such as LLaMA-Mesh \citep{wang2024llamamesh} together with human Likert-scale ratings across multiple dimensions of shape quality. By contrast, the reward signal we use is extracted from a deliberately simple pipeline: for preference elicitation a user is shown between two and six visualised geometries and asked only to select the highest- and lowest-quality samples according to whatever criteria they consider important. A practical appeal of this setup is that the reward model can be learned from simple preference pairs over the implicit $\sigma$ field alone, without further information, reaching $91\%$ accuracy on held-out within-distribution pairs (Section~\ref{sec:results:reward}). A broader implication is that a pretrained generator can then be fine-tuned on its own learned distribution so as to alter one part of its representation, the 3D geometry, while leaving 2D appearance qualitatively similar at bounded cost.

\section{Method}\label{sec:method}

\subsection{Extracting shape features from a NeRF}\label{sec:method:features}

A NeRF \citep{mildenhall2020nerf} is a mapping from 3D spatial coordinates $x, y, z$ and viewing direction $\theta, \Phi$ to colour $R, G, B \in [0,1]$ and density $\sigma \in \mathbb{R}$:
\begin{equation}
F_{\Theta} \!: (x, y, z, \theta, \Phi) \to (R, G, B, \sigma).
\label{eq:nerf}
\end{equation}
Images are generated from this field via volume rendering, encoding realistic features such as view-dependent specularities and partial opacity of regions with non-zero density. Generative NeRFs \citep{schwarz2020graf} extend Equation~\eqref{eq:nerf} into a distribution of radiance fields for images of a single object category such as faces. Persistent defects in the recovered 3D geometry remain, however \citep{chan2022eg3d,shi2022geod,karras2019ffhq,skorokhodov2022epigraf}. Although alternative representations such as signed-distance fields or explicit meshes offer stronger geometric priors \citep{li2025craftsman3d,ye2025hi3dgen,xiang2024trellis}, all approaches exhibit issues due to the ill-posed problem of inferring 3D shape from 2D projections. Our contribution focuses on fine-tuning the learned 3D geometry with human feedback without needing a mesh prior.

Rather than extracting an explicit mesh from the density volume - which a differentiable iso-surface model such as DMTet \citep{shen2021dmtet} could in principle do, at additional computational cost - our approach works directly on the sigma feature maps, encoding the implicit geometry into a reward score with a 3D~U-Net ResNet backbone applied to $\sigma$ sampled over the scene volume. The resulting reward model is inexpensive: both training and fine-tuning take approximately $5$--$10$ hours on a single RTX~4090, depending on the 3D representation used.

\paragraph{Using density $\sigma$.}
Of the tuple $(R, G, B, \sigma)$ returned by the field, only $\sigma$ contains the shape information of the radiance field; the colour channels should change as the camera is moved. To simplify our approach we extract shape features conditional on the fixed view angles $\theta_{c}, \Phi_{c}$ termed the canonical view: $\xi_{c} = (\theta_{c}, \Phi_{c})$ corresponds to viewing poses occurring in the middle of the viewing-pose distribution of FFHQ \citep{karras2019ffhq}, where most images are taken with the subject facing the camera. From the canonical-view radiance field $F_{\Theta}(x, y, z, \theta\!=\!\theta_{c}, \Phi\!=\!\Phi_{c})$ we consider three differentiable 3D representations: the depth map, point cloud, and sigma field.

\paragraph{Depth map.}
The depth map is an image of dimension $H \!\times\! W$ where each pixel records the expected stopping depth of light along its ray. The depth value $D(\mathbf{r})$ along a ray $\mathbf{r}$ is defined through the transmittance $T(t)$ as
\begin{equation}
D(\mathbf{r}) = \int_{t_{n}}^{t_{f}} T(t)\,\sigma(\mathbf{r}(t))\,t\,\mathrm{d}t,
\quad
T(t) = \exp\!\Big(\!-\!\!\int_{0}^{t} \sigma(\mathbf{r}(s))\,\mathrm{d}s\Big),
\label{eq:depth}
\end{equation}
where $\sigma(\mathbf{r}(t))$ is the density of the radiance field at $\mathbf{r}(t)$. We use the quadrature approximation of \citet{mildenhall2020nerf} and estimate depths at resolution $H \!=\! W \!=\! 128$. We consider both the single canonical-view depth map and a triple-view variant that additionally renders the two off-canonical views at $\pm 60^\circ$ yaw, giving the reward model multi-view geometric context.

\paragraph{Point cloud.}
The point cloud representation is computed from the depth map by converting each pixel depth to its $(x,y,z)$ coordinate in world space. For a depth-map camera pose $\xi$, a rendered image of resolution $H \!\times\! W$ entails a collection of rays starting from the pinhole centre $\vec{r}_{0}$. Denoting the direction of the ray at pixel $(h,w)$ by $\vec{r}_{d}(h,w)$, each depth-map pixel has a corresponding ray $\vec{r} = \vec{r}_{0} + t \times \vec{r}_{d}(h,w)$, $t \in \mathbb{R}^{+}$. Writing $d(h,w)$ for the depth $D(\mathbf{r})$ at $(h,w)$, the point-cloud representation places one point per pixel:
\begin{equation}
p_{h,w} = \vec{r}_{0} + d(h,w) \times \vec{r}_{d}(h,w).
\label{eq:pointcloud}
\end{equation}
Repeating this for all $H \!\times\! W = 16{,}384$ pixels yields a point cloud of $16{,}384$ points.

\begin{figure}[htbp]
  \centering
  \includegraphics[width=0.95\linewidth]{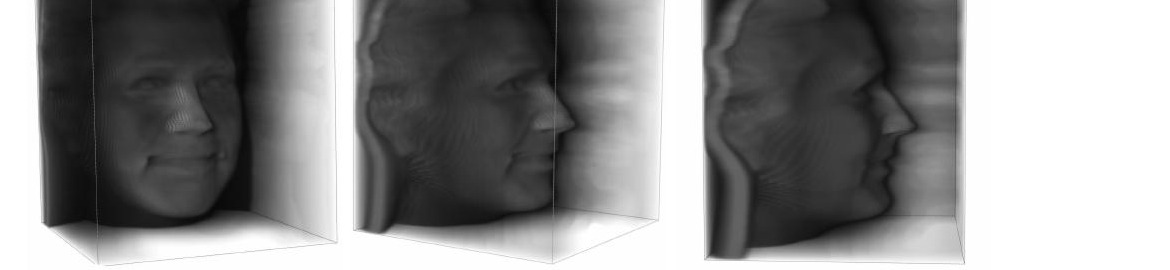}
  \caption{Sigma field $\sigma_{XYZ}$ visualised from three rotated view angles. Each panel shows the density field of a single EG3D-FFHQ sample, rendered as a volume with the cube outline visible.}
  \label{fig:sigma_volume}
\end{figure}

\paragraph{Sigma field.}
The sigma field representation consists of $\sigma$ values extracted over a fixed 3D coordinate grid in scene space, denoted $\sigma_{XYZ}$. While $\sigma_{XYZ}$ contains geometric information of the scene volume, the other two representations encode information about the estimated surface of the geometry only. Figure~\ref{fig:sigma_volume} depicts the $\sigma_{XYZ}$ representation, where pixel intensities correspond to the amount of accumulated sigma weight along the corresponding ray.

\paragraph{Reward-input slab and normalisation.}
For the reward model we do not feed the full $256^{3}$ sigma cube. Instead, the sampled cube is cropped to a fixed frontal face slab inherited from the EG3D training pipeline: at resolution $256$, the kept region is $X[64{:}192]$, $Y[64{:}205]$, $Z[102{:}231]$, giving a tensor of shape $128 \times 141 \times 129$. Equivalently, this trims $25\%$ from the left and right, $25\%$ from the bottom, $20\%$ from the top, $40\%$ from the rear, and $10\%$ from the front of the volume. The crop is a memory-driven compromise rather than a purely semantic one: geometric defects become more apparent as $\sigma$ is sampled at higher resolution, so finer feedback is desirable, but extracting the full $256^{3}$ cube (let alone $512^{3}$) and training the volumetric reward model on it exceeds the $24$\,GB of an RTX~4090. Cropping to the frontal face slab retains the facial regions that carry the preference signal while keeping the input resolution as high as the hardware allows, and discards mostly empty background outside the canonical-view head shell.

\begin{figure}[htbp]
  \centering
  \includegraphics[width=0.95\linewidth]{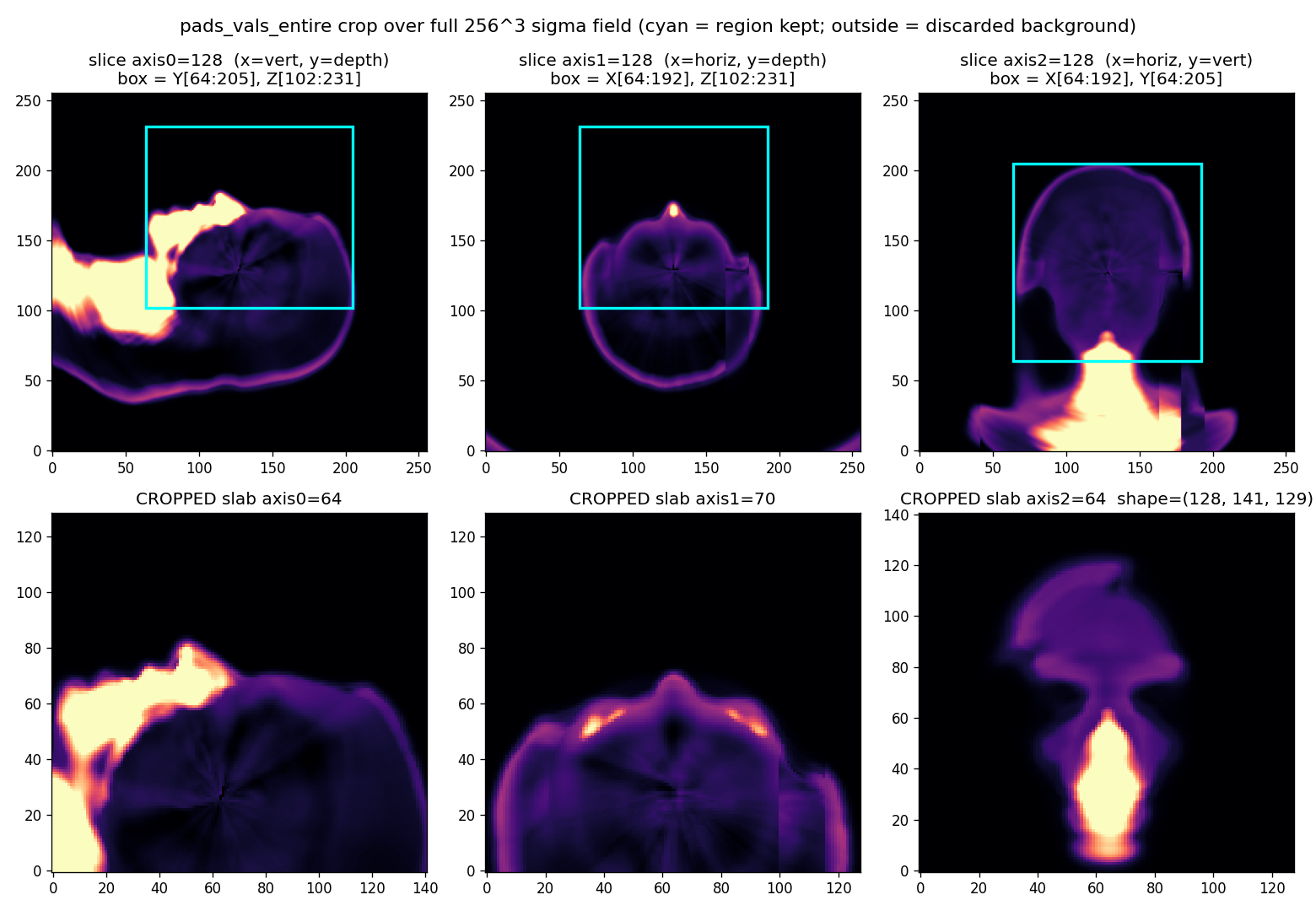}
  \caption{Reward-input slab used for $\sigma_{XYZ}$ scoring. The top row shows the crop box inside the full $256^{3}$ EG3D sigma cube on three orthogonal slices; the bottom row shows the resulting cropped tensor of shape $128 \times 141 \times 129$. After cropping, each slab is independently rescaled to $[0,100]$ by
  \texttt{normalise\_sigma\_self}.}
  \label{fig:reward_slab_crop}
\end{figure}

Unless stated otherwise, each cropped slab is then transformed by the per-cube map used in the codebase as \texttt{normalise\_sigma\_self},
\[
x \mapsto 100 \times \frac{x - \min(x)}{\max(x) - \min(x)},
\]
so every slab is min-max scaled to the common range $[0,100]$ before reward scoring. Figure~\ref{fig:reward_slab_crop} shows the retained slab relative to the full $256^{3}$ cube and the resulting cropped tensor.

\subsection{Learning a model of 3D shape quality from preference pairs}\label{sec:method:reward}

Prior work has shown that a model of 3D shape quality can be learned for explicit meshes, often with the aid of language prompts, and that such a model agrees with human preference. A mesh is a stronger geometric prior than a radiance field, in which the geometry is only implicit: a mesh surface is explicit, so defects such as an open (unbounded) surface can be characterised and repaired directly, and many methods exist for reconstructing a closed surface from a set of candidate boundary points or point clouds. Arbitrarily closing or re-connecting a surface does not by itself yield plausible geometry, however - it may introduce intersecting or spiky faces - and it is here that a human-preference reward model adds value, favouring regular, continuous, high-quality surfaces. Mesh-based fine-tuning is moreover inherently discrete, acting on a finite vertex graph. Our findings suggest that it can instead be advantageous to keep the representation implicit, avoiding the conversion of a radiance field to a mesh and back - a round trip that has no canonical inverse and would require approximation. Although human preferences for 3D shape quality are often immediately apparent when inspecting visualised radiance-field geometry, distilling them into a learnable quality-scoring module $r_{\theta}$ is non-trivial. The pipeline we use is nonetheless simple: from a dataset of ranked sampled shapes, a model is trained to predict pairwise preferences from shape features, and this model is then used to fine-tune EG3D.

\subsubsection{Creating a dataset of human preferences}\label{sec:method:dataset}

\paragraph{Extracting preference data.}
We collect user preferences by synthesising radiance fields $(R,G,B,\sigma) = G(z)$ from the pretrained generator $G$, visualising the geometry from the $\sigma$ values using the Marching Cubes algorithm \citep{lorensen1987marchingcubes}, and asking human respondents to rank the visualised geometries. An initial attempt at multi-respondent triplet ranking $[x_{1}, x_{2}, x_{3}]$ did not converge to a stable preference. We instead extract preferences from the primary researcher of the study via a questionnaire that elicits a ranking over batches of 3D geometries. Each question contains between two and six examples; while this could provide many pairwise comparisons under combinatorial scaling, the data is reduced for training a performant reward model. The reduced sample extracts the highest-ranked example $x_{w}$ and the lowest-ranked example $x_{l}$ from each batch, discarding all other ranked examples. This yields $n = 4{,}346$ preferred/dispreferred training pairs $[x_{w} \!\succ\! x_{l}]$.

\paragraph{Augmenting preference data.}
In each batch the researcher selects the highest-quality sample $x_{w}$ and the lowest-quality sample $x_{l}$. One issue with this data is that the winning sample $x_{w}$ can itself still contain defects, or be of poor quality. During training, the learned preference of the reward model encodes such defects as preferable, which might push the fine-tuned EG3D geometries towards emulating such poor-quality features. We address this by augmenting each ranking batch with a single high-quality sample drawn from the centre of the GAN's latent space, which we refer to as $x_{HQ}$. While such samples lack diversity, they are almost certain to be of much better quality than either $x_{w}$ or $x_{l}$. The final training batch contains three samples such that $x_{HQ} \!\succ\! x_{w} \!\succ\! x_{l}$. After augmentation there are $n = 4{,}346$ ranking batches with three examples in each. We split into train/validation/test partitions with proportions $0.7 / 0.15 / 0.15$ ($3{,}042 / 652 / 652$ batches). Even when the $x_{HQ}$ anchor is removed at test time, the reward model still discriminates the harder regular-versus-regular pairs at $91\%$ accuracy (Section~\ref{sec:results:reward}), so the learned signal is not merely detecting the conspicuous high-quality anchor.

\subsubsection{Reward-model architecture and training}\label{sec:method:arch}

\begin{figure}[htbp]
  \centering
  \includegraphics[width=0.9\linewidth]{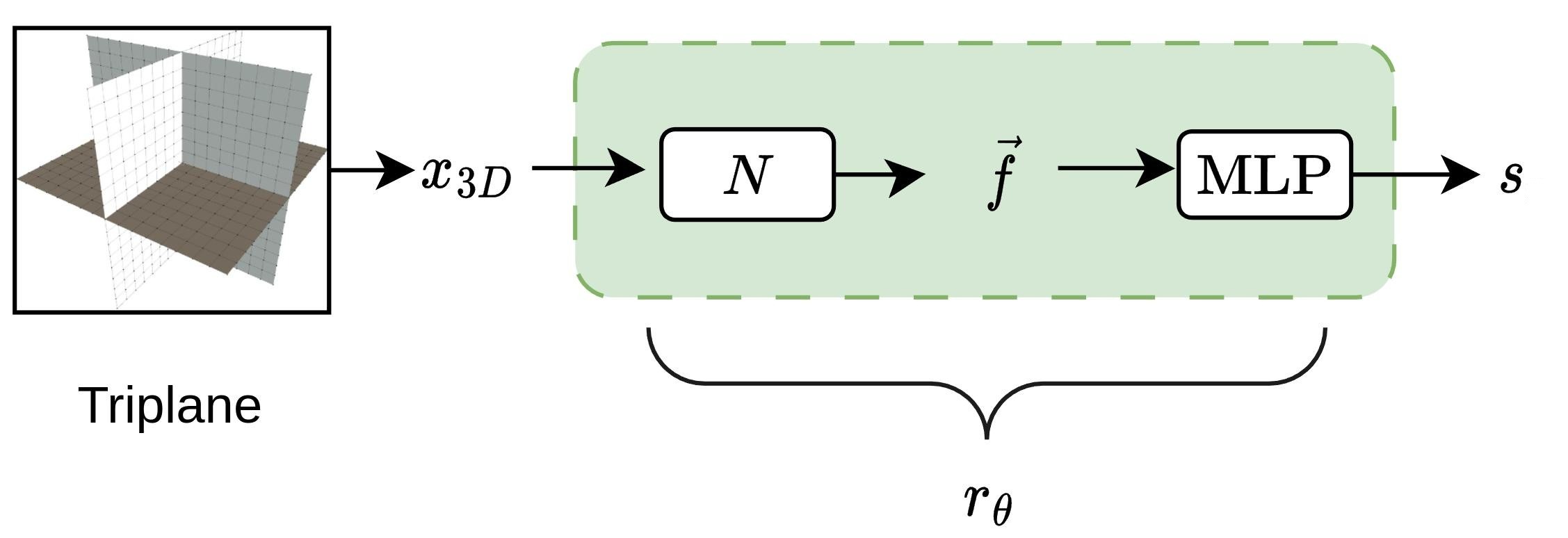}
  \caption{The reward model $r_{\theta}$ predicts a quality score $s$ from 3D representation $x_{3D}$. The module $N$ is a domain-specific feature extractor mapping $x_{3D}$ to a global feature $\vec{f}$. An MLP decodes $\vec{f}$ into the quality score.}
  \label{fig:reward_arch}
\end{figure}

\paragraph{Reward-model architecture.}
The reward-model architecture, depicted in Figure~\ref{fig:reward_arch}, enables experimentation with multiple feature extractors. The first sub-module $N$, chosen based on the 3D representation, maps $x_{3D}$ (a depth map, point cloud, or sigma field) into a global feature vector $\bar{f}$. The second sub-module is an MLP that maps $\bar{f}$ into the quality score $s$. We experiment with the following sub-modules $N$. For depth maps we consider ResNet-50 \citep{he2016resnet}, VGGFace \citep{parkhi2015vggface} and VGG-4096 \citep{simonyan2014vgg}. For point clouds we consider PointNet \citep{qi2017pointnet}, PointNet++ \citep{qi2017pointnet++}, and CurveNet \citep{xiang2021curvenet}. For the sigma field we consider 3D U-Net variants \citep{cicek2016unet3d,hu2018senet,wolny2020plant3d,toubal2020aipr}, of which a squeeze-and-excitation residual variant (ResNet-SE-3D-UNet) performs best and is used throughout. Full architecture specifications and training-configuration files are provided in the released code repository.

\paragraph{Reward-model training loss.}
The reward-model training loss $\mathcal{L}_{\theta} = \mathcal{L}_{w}$ is the pairwise prediction loss in Equation~\eqref{eq:lw}, which encourages $r_{\theta}$ to predict the winning example $x_{w}$ over ranked pairs in a minibatch of size $K$. It is derived from Equation~(1) of \citet{ouyang2022instructgpt}, modified to remove prompt conditioning:
\begin{equation}
\mathcal{L}_{w} = -\frac{1}{\binom{K}{2}} \mathbb{E}_{(x_{w}, x_{l}) \sim D}
   \!\left[\log\!\big(S(r_{\theta}(x_{w}) - r_{\theta}(x_{l}))\big)\right],
\label{eq:lw}
\end{equation}
where $S$ denotes the sigmoid function in ~\ref{eq:lw}. The same objective is used for all 3D representations (depth map, point cloud, and $\sigma_{XYZ}$). For the sigma-field reward model only, we additionally use an auxiliary reconstruction loss on the 3D U-Net output: the network reconstructs the input cropped sigma slab, this reconstruction is normalised back to the slab's input scale, and an $L^{1}$ penalty is applied with weight $10^{-2}$. This stabilises the volumetric feature extractor while the pairwise preference objective remains the main supervision signal.

\paragraph{Hyperparameters.}
Reward models are trained using Adam \citep{kingma2014adam} with a learning rate of $10^{-5}$ and weight-decay parameter $10^{-4}$. The batch size varies based on setting: for depth maps, point clouds, and $\sigma_{XYZ}$ the batch sizes were $8$, $2$, and $1$ respectively. Reward models were trained for a maximum of $10$ epochs, with early stopping if the validation loss was not improved over three epochs. Experiments were carried out on a single Nvidia RTX 4090.

\subsubsection{Fine-tuning EG3D geometry}\label{sec:method:finetune}

\begin{figure}[htbp]
  \centering
  \includegraphics[width=0.95\linewidth]{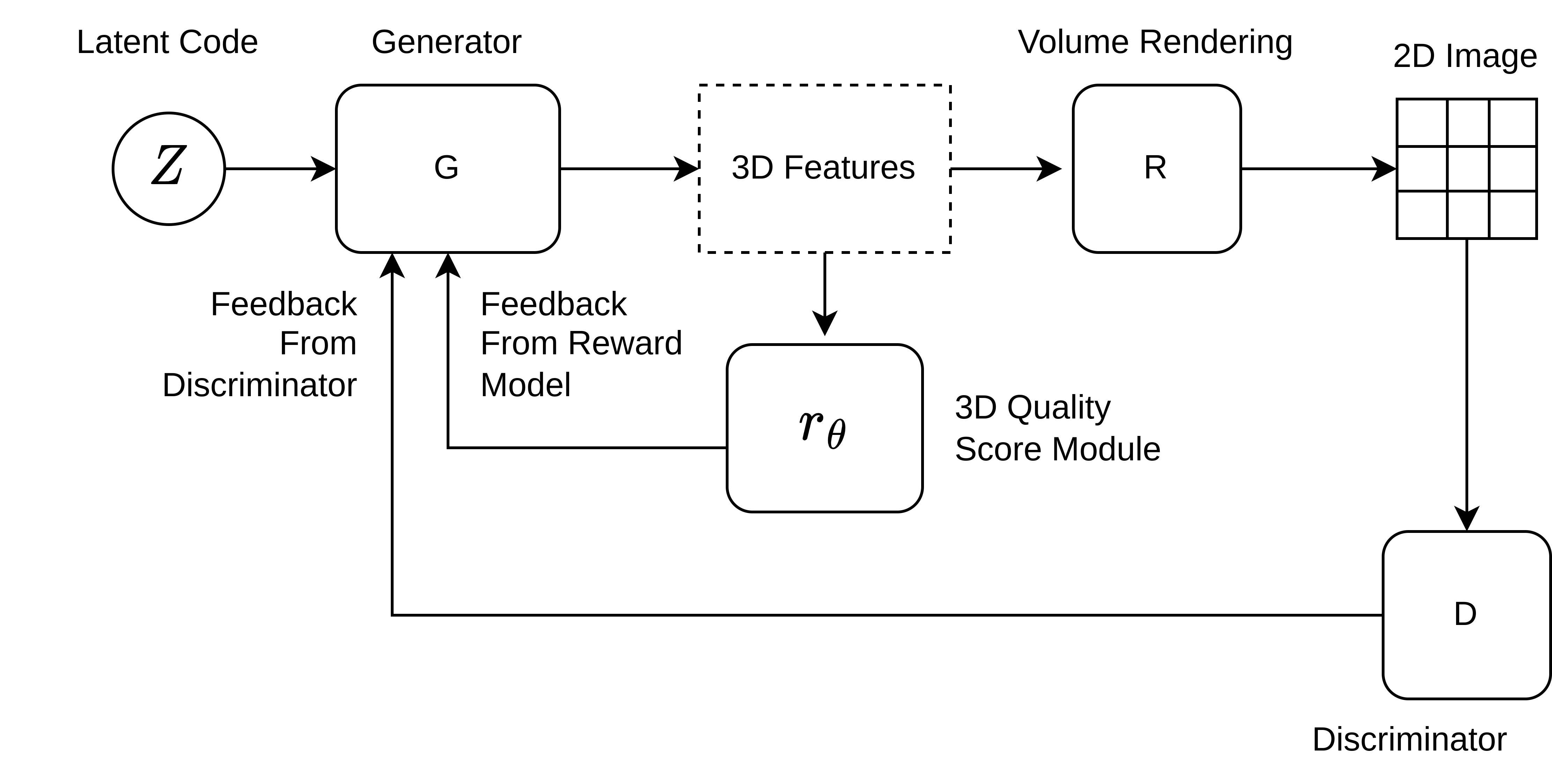}
  \caption{Modification of the generator update step in the GAN loss. The reward model $r_{\theta}$ scores the 3D feature volume produced by the generator $G$, and the resulting reward signal is fed back to $G$ alongside the discriminator's feedback.}
  \label{fig:gan_modification}
\end{figure}

Given the reward model $r_{\theta}$, our goal is to fine-tune the generator to improve the geometry while limiting degradation in 2D image quality. Our experiments suggest that this task is best accomplished with the original GAN loop, training both the discriminator and the generator for a small number of additional steps. The joint incorporation of feedback from $r_{\theta}$ and the discriminator is depicted in Figure~\ref{fig:gan_modification}. We leave the discriminator loss unchanged,
\begin{equation}
\mathcal{L}_{D}
   = -\tfrac{1}{2}\mathbb{E}_{x \sim p_{s}}\!\log D(x)
   - \tfrac{1}{2}\mathbb{E}_{z \sim p_{z}}\!\log\!\big(1 - D(G(z))\big)
   + \gamma_{R_{1}} \!\times\! R_{1},
\label{eq:ld}
\end{equation}
where the $R_{1}$ penalty regularises the discriminator \citep{mescheder2018r1}.

\paragraph{Generator loss $\mathcal{L}_{G}$.}
We modify the generator loss to incorporate feedback from the reward model, appending two extra terms: the reward loss $\mathcal{L}_{r}$ and 3D consistency loss $\mathcal{L}_{c}$:
\begin{equation}
\mathcal{L}_{G}
  = \underbrace{-\tfrac{1}{2}\mathbb{E}_{z \sim p_{z}} \log D(G(z))}_{\text{original GAN loss}}
  + \underbrace{\lambda_{r}\, \mathcal{L}_{r}}_{\text{reward loss}}
  + \underbrace{\lambda_{c}\, \mathcal{L}_{c}}_{\text{consistency loss}}.
\label{eq:lg}
\end{equation}
Unless otherwise specified, $\lambda_{c} = 10^{-2}$ and $\lambda_{r} = 10$.

\paragraph{Reward loss $\mathcal{L}_{r}$.}
The reward loss uses the quality score $s = r_{\theta}(x_{3D})$ computed by the reward model, averaged over the generator samples in the minibatch. The function $f_{r}$ clamps this raw score to $[-10, 10]$, and the result is negated, $\mathcal{L}_{r} = -f_{r}(s)$, so that minimising $\mathcal{L}_{G}$ maximises the reward. A side effect of the clamp is that once a sample's score reaches the bound its reward gradient vanishes, so examples that already score a high reward are no longer pushed in the direction of a further reward increase; the reward signal therefore acts most strongly on the lower-scoring geometries, while the discriminator and consistency terms preserve image quality and identity. We found the clamp to be an experimental but necessary addition for training stability: by bounding the per-step reward it keeps the dynamics stable around the existing GAN training loop and prevents runaway reward values that would otherwise destabilise the joint generator--discriminator update, at the cost of no longer optimising samples already judged high-quality. The clamp also bounds the gradient regardless of where the reward model's output distribution sits, so the raw score is used directly, without re-centring or rescaling.

\paragraph{Consistency loss $\mathcal{L}_{c}$.}
The consistency loss prevents the geometry from the fine-tuned generator $G_{\text{new}}$ from diverging too far from that of the original $G_{\text{old}}$. Density values are extracted from $G$ over a grid of resolution $64^{3}$ along $X{-}Y{-}Z$ scene coordinates, denoted $\sigma^{64} \!\circ G$. The consistency loss is the $L^{1}$ distance between the new $\sigma$ values and those drawn from the pretrained EG3D, $G_{\text{old}}$:
\begin{equation}
\mathcal{L}_{c}
  = \mathbb{E}_{z \sim p_{z}}
     L^{1}\!\big[\sigma^{64} \!\circ G_{\text{new}}^{z}, \;
                 \sigma^{64} \!\circ G_{\text{old}}^{z}\big].
\label{eq:lc}
\end{equation}

\paragraph{Hyperparameters.}
From the original EG3D training pipeline, the batch size is decreased from $b_{s} \!=\! 32$ to $b_{s} \!=\! 16$, and $\gamma_{R_{1}}$ is increased from $1$ to $20$. The EG3D density regularisation is retained at its default strength, distinct from our consistency loss $\mathcal{L}_{c}$. Although $\lambda_{c}=10^{-2}$ is much smaller than $\lambda_{r}=10$, it still provides a persistent pull toward the pretrained density field because it is applied on every update step over a full sampled sigma grid. All remaining hyperparameter choices are unchanged; see \citet{chan2022eg3d}. Fine-tuning comprises $20$ kimg ($\approx\!20{,}000$ images on the original FFHQ dataset that was resynthesised according to the original EG3D specifications).

\section{Experiments and Results}\label{sec:results}

Results are organised into two parts: reward-model training and evaluation (Section~\ref{sec:results:reward}), and the result of using the reward model to fine-tune the generator (Section~\ref{sec:results:finetune}).

\subsection{Reward-model training}\label{sec:results:reward}

\paragraph{Reward-model performance and evaluation.}
Test accuracy - the fraction of held-out preference pairs predicted correctly - clearly favours the sigma-field ($\sigma_{XYZ}$) representation on the hard within-distribution comparisons that matter for fine-tuning, where it reaches $0.91$, against $0.74$ for the strongest image-derived reward - the triple-view depth map - while the single-view depth map and all three point-cloud backbones collapse to chance ($\approx 0.50$) (Table~\ref{tab:repaccuracy}, \emph{regular only}). The $\sigma_{XYZ}$ reward is ahead of every image-derived representation - mirroring its relative effectiveness during fine-tuning, where only the $\sigma_{XYZ}$ reward induces favourable geometry change. Each ranking the model is scored on contains a high-quality anchor - a low-truncation sample that is conspicuously cleaner than the rest - so a substantial fraction of the test pairs are easy. Including this anchor raises the apparent accuracy to $0.97$ for $\sigma_{XYZ}$ and $0.91$ for the triple-view depth map; the single-view depth map and PointNet detect the anchor well ($0.83$), whereas PointNet++ and CurveNet do not ($\approx 0.50$) (Table~\ref{tab:repaccuracy}, \emph{all pairs}). Removing the anchor and scoring only the harder within-distribution comparisons sharpens the contrast. This harder regime is precisely the one that matters for fine-tuning: because EG3D is sampled \emph{without} truncation during fine-tuning, the generator's outputs lie in the untruncated part of the distribution rather than near the high-quality anchor. The $\sigma_{XYZ}$ reward's ability to discriminate \emph{within} that distribution, where the point-cloud and single-view depth-map rewards collapse to chance and even the triple-view depth map, though stronger, still trails $\sigma_{XYZ}$, helps account for its effectiveness as a fine-tuning signal.

\begin{table}[t]
\caption{Test accuracy of $r_{\theta}$ by 3D representation: the fraction of held-out preference pairs predicted correctly, on a common held-out split of the labelled ranking data ($652$ ranking questions), each model using its own input pipeline. \emph{All pairs} ($1{,}956$ pairs) includes the high-quality low-truncation anchor present in each ranking; \emph{regular only} ($652$ pairs) removes that anchor, leaving the harder within-distribution comparisons that match the untruncated regime used during fine-tuning.}
\label{tab:repaccuracy}
\centering
\begin{tabular*}{\tblwidth}{@{}LLLL@{}}
\toprule
Representation & Backbone (input) & All pairs & Regular only \\
\midrule
Sigma field & ResNet-SE-3D-UNet ($256^3$ slab) & 0.97 & 0.91 \\
Depth map    & ResNet-50 (single canonical view) & 0.83 & 0.50 \\
Depth map    & ResNet-50 (triple view, $\pm 60^\circ$ yaw) & 0.91 & 0.74 \\
Point cloud  & PointNet ($16{,}384\!\to\!2{,}048$ pts) & 0.83 & 0.50 \\
Point cloud  & PointNet++ ($16{,}384\!\to\!2{,}048$ pts) & 0.50 & 0.50 \\
Point cloud  & CurveNet ($16{,}384\!\to\!2{,}048$ pts) & 0.51 & 0.51 \\
\bottomrule
\end{tabular*}
\end{table}

\paragraph{Which 3D representation works best?}
The sigma field $\sigma_{XYZ}$ is the most effective representation on held-out test accuracy (Table~\ref{tab:repaccuracy}): $0.91$ on the hard within-distribution pairs that matter for fine-tuning, against $0.74$ for the best depth-map model (the triple-view variant) and $\approx 0.50$ for every point-cloud backbone. Test accuracy alone is sufficient to select a representation, and the ordering it gives is borne out during fine-tuning, where only the $\sigma_{XYZ}$ reward induces a favourable change in 3D geometry; geometries ranked highly by the depth-map or point-cloud rewards tended to retain persistent surface defects - and even the stronger triple-view depth-map reward still ranks some clearly defective geometries (e.g.\ over-sharp noses) highly. Our best $\sigma_{XYZ}$ model with modest computational cost uses a ResNet-SE-3D-UNet architecture \citep{cicek2016unet3d,hu2018senet,wolny2020plant3d,toubal2020aipr}; a comparison of geometries induced by the best-performing backbone of each representation is shown in Figure~\ref{fig:geom_reward_compare}.

This result suggests that the sigma density volume carries information useful both for learning a 3D quality model and for fine-tuning a generator that was never trained on 3D data directly. A NeRF is a weaker geometric prior than a mesh: a mesh has an explicit surface and is amenable to methods that directly measure the continuity or regularity of that surface, whereas the implicit density field makes it less obvious how to enforce geometric regularity without an explicit mesh or an intermediate marching-cubes step. Our reward model instead operates on the scene volume's $\sigma$ density - including empty space - rather than on a known surface, so the quality-scoring module is in this sense surface-agnostic and learned with comparatively weak supervision. Because $\sigma$ rises sharply at the ray-termination point, the volume nonetheless carries enough information to localise the surface, and the resulting reward improves the geometry (viewed via marching cubes) while minimally altering the appearance in RGB space.

\subsection{Fine-tuning EG3D shapes}\label{sec:results:finetune}

\begin{figure}[htbp]
  \centering
  \includegraphics[width=0.95\linewidth]{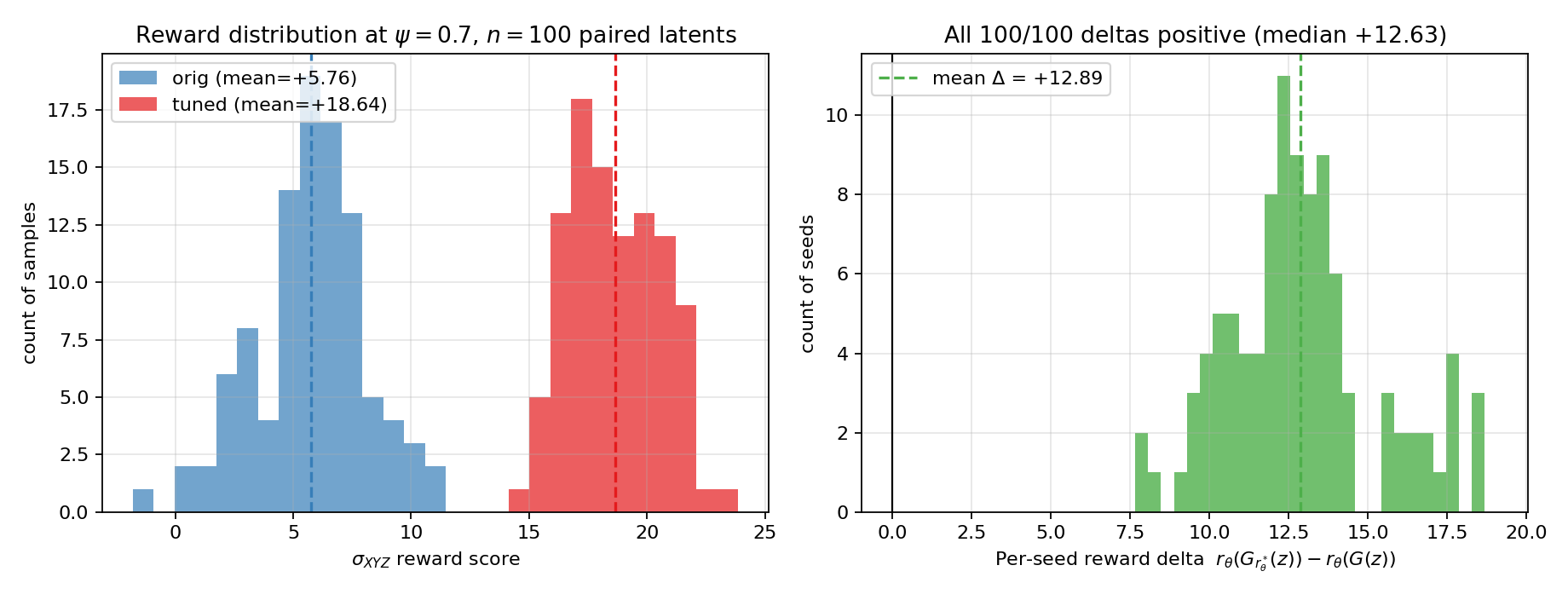}
  \caption{Change in $\sigma_{XYZ}$ reward distribution after fine-tuning, on $100$ paired latent codes at truncation $\psi=0.7$. \emph{Left:} histograms of reward scores before (orig, blue) and after (tuned, red) fine-tuning, with dashed lines marking the means. \emph{Right:} distribution of per-seed deltas $r_{\theta}(G_{r_{\theta}^{*}}(z)) -r_{\theta}(G(z))$. All $100/100$ deltas are positive with mean $+12.89$.}
  \label{fig:reward_hist}
\end{figure}

\begin{figure}[htbp]
  \centering
  \includegraphics[width=0.49\linewidth]{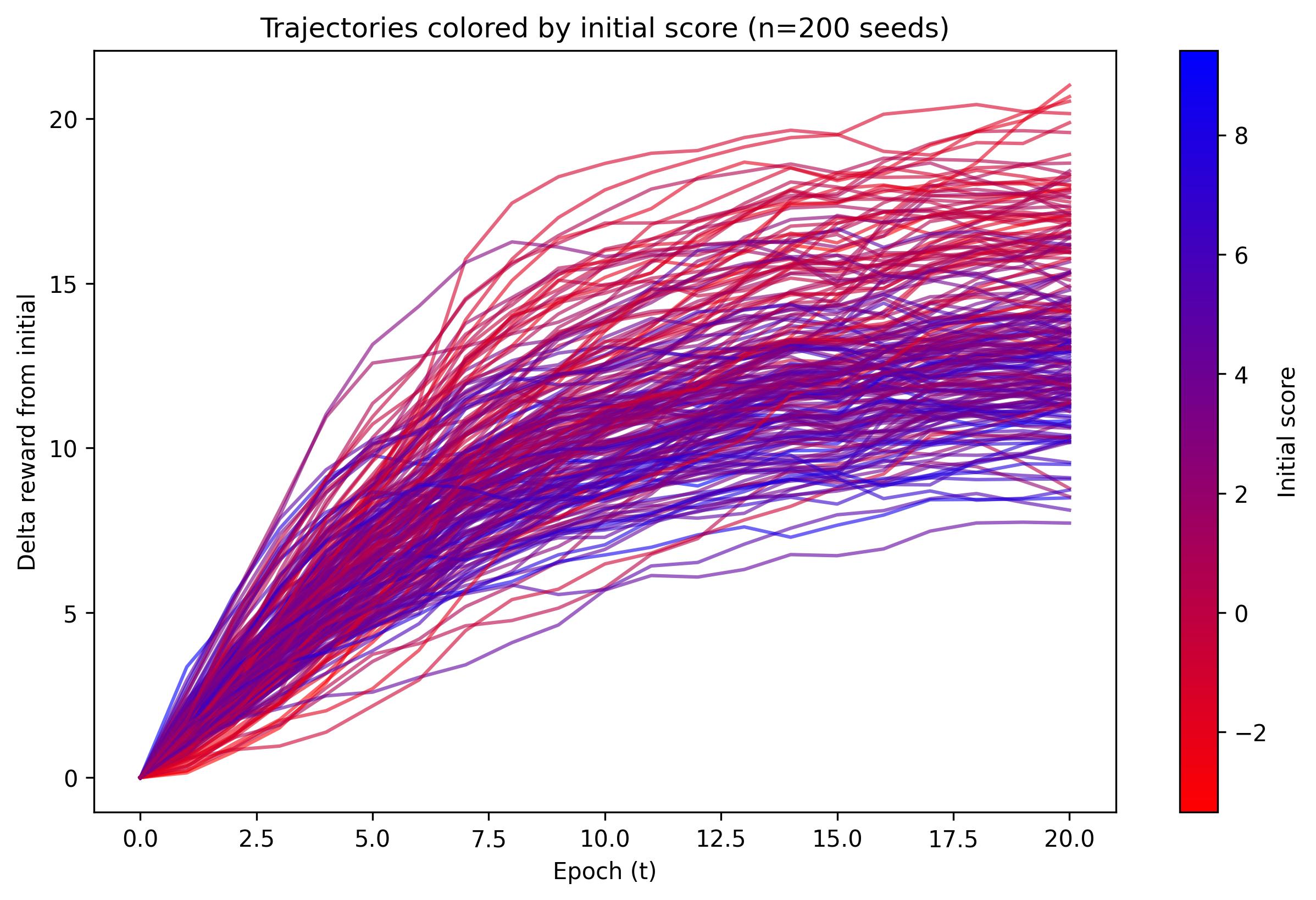}\hfill
  \includegraphics[width=0.49\linewidth]{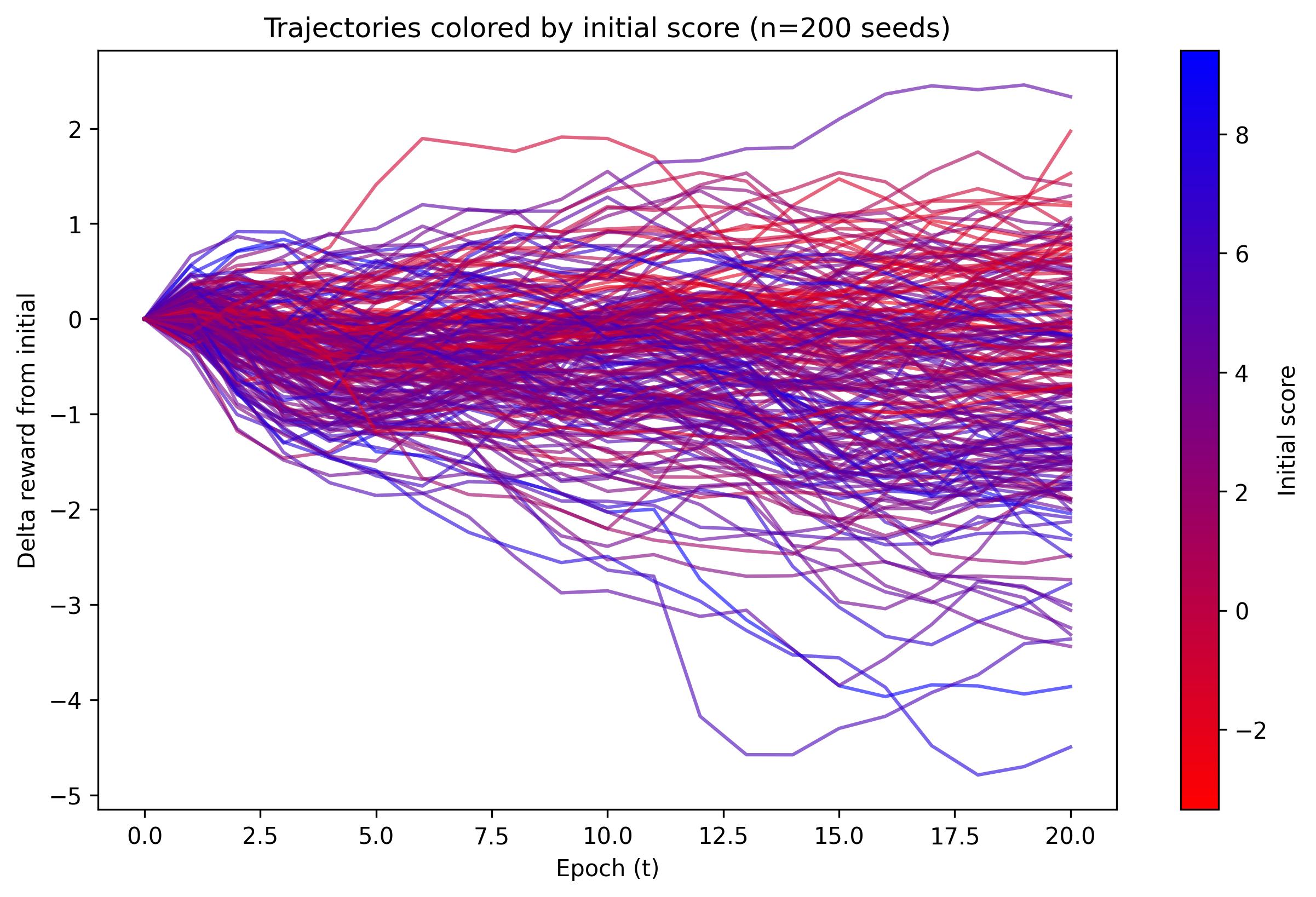}
  \caption{Per-seed $\sigma_{XYZ}$ reward trajectories during fine-tuning, for $200$ fixed latent codes, each line coloured by its initial reward score. \emph{Left:} with the reward loss ($\lambda_{r} = 10$) the reward rises and saturates for essentially every seed. \emph{Right:} the matched no-reward control ($\lambda_{r} = 0$) shows no systematic reward change. The mean per-seed reward increase is large under the reward loss and approximately zero for the control, confirming that the geometry improvement is driven by the reward signal rather than by continued GAN training.}
  \label{fig:reward_traj}
\end{figure}

\begin{figure}[htbp]
  \centering
  \begin{minipage}{0.49\linewidth}\centering
    \includegraphics[width=\linewidth]{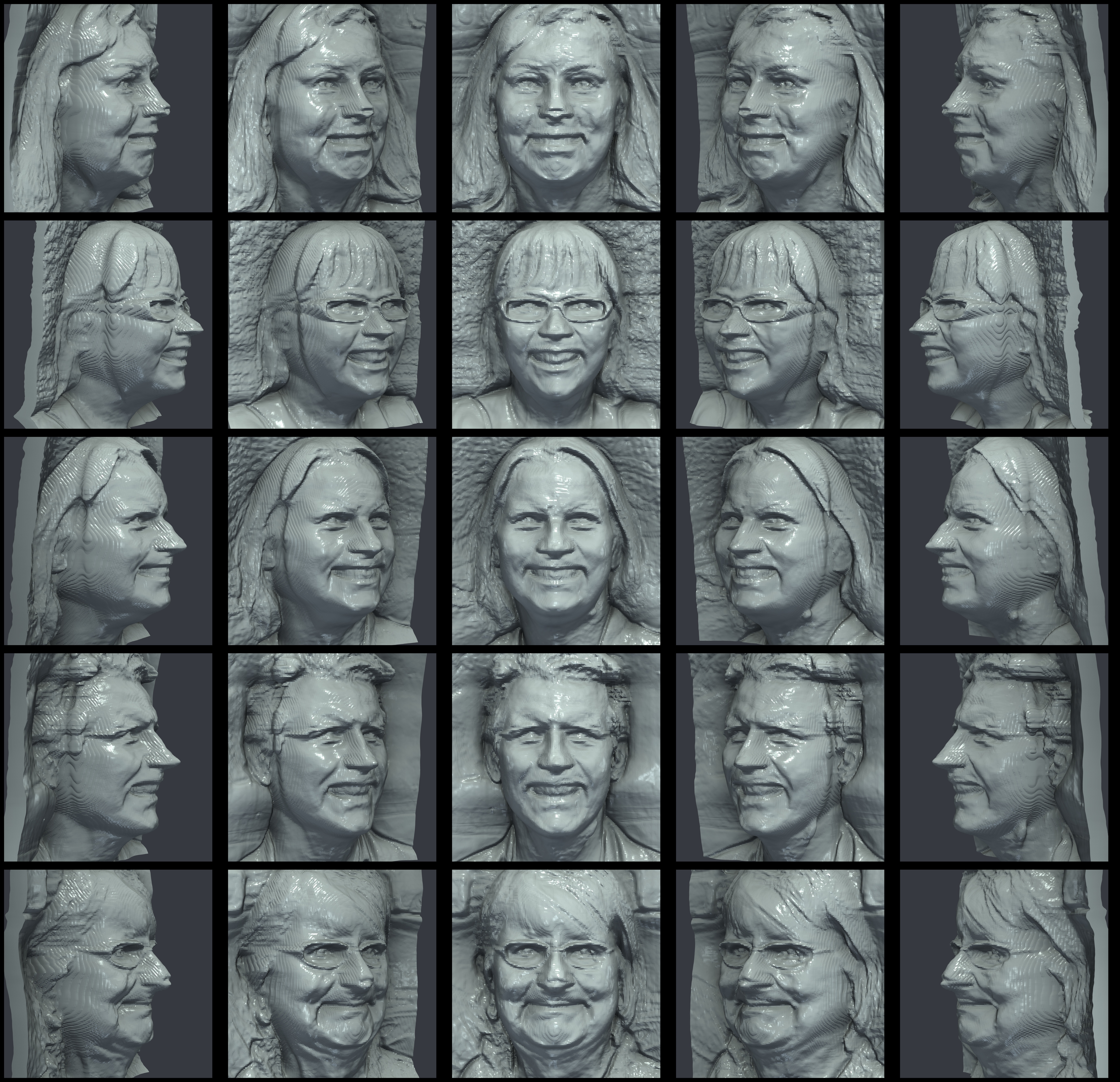}\\[1pt]{\small (a) no reward ($\lambda_r=0$, control)}
  \end{minipage}\hfill
  \begin{minipage}{0.49\linewidth}\centering
    \includegraphics[width=\linewidth]{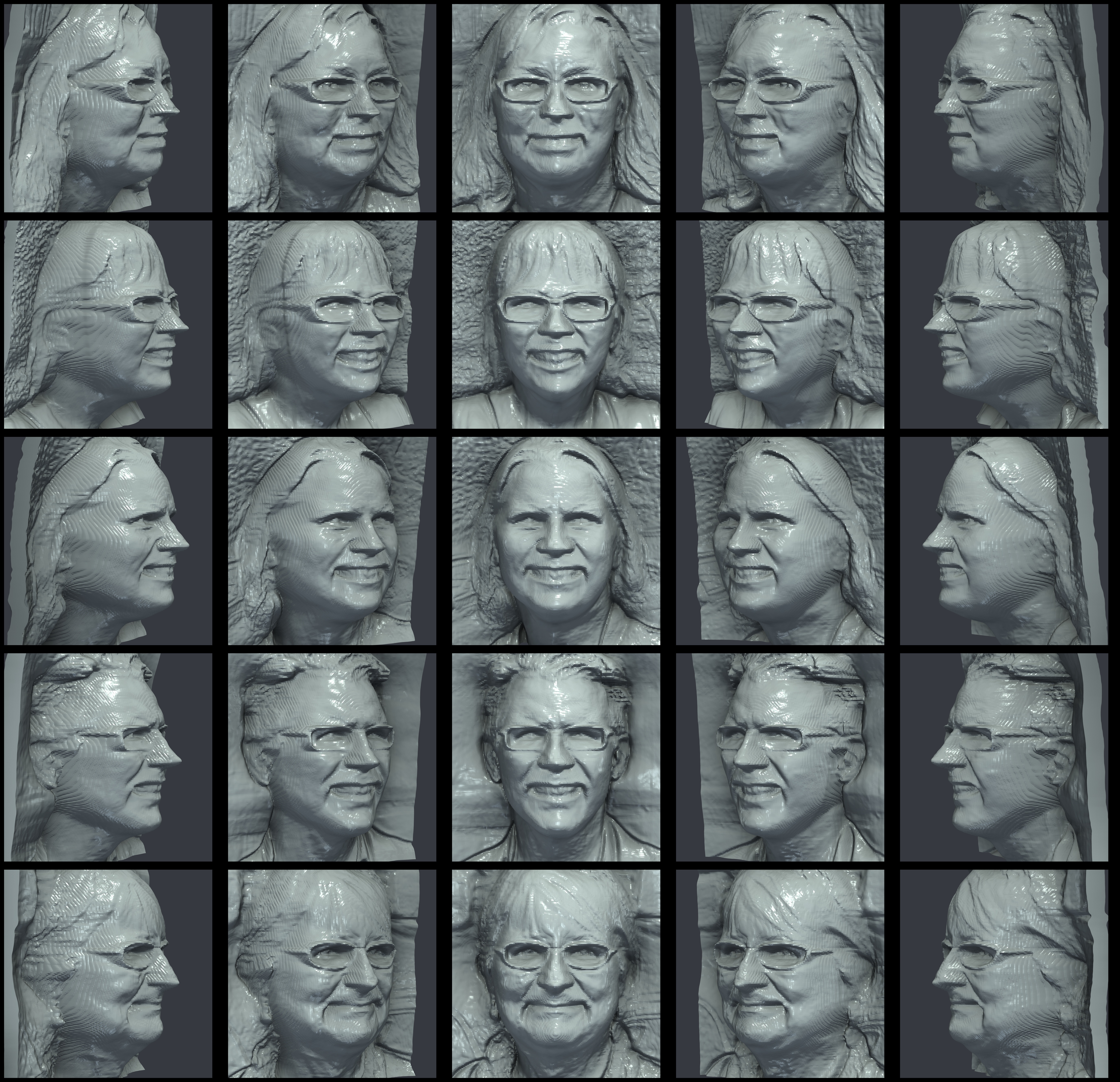}\\[1pt]{\small (b) $\sigma_{XYZ}$ reward}
  \end{minipage}
  \\[4pt]
  \begin{minipage}{0.49\linewidth}\centering
    \includegraphics[width=\linewidth]{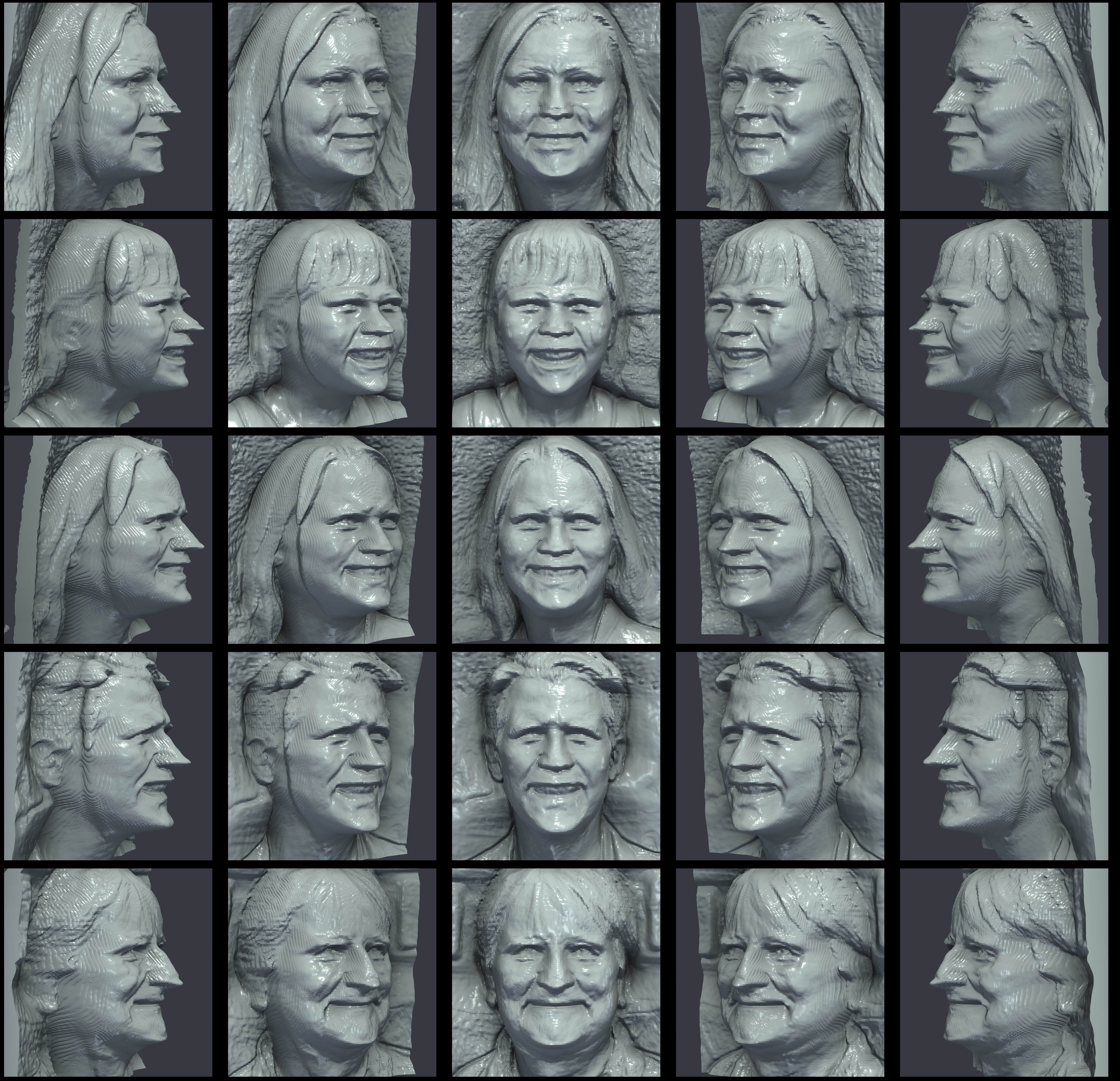}\\[1pt]{\small (c) triple-view depth-map reward}
  \end{minipage}\hfill
  \begin{minipage}{0.49\linewidth}\centering
    \includegraphics[width=\linewidth]{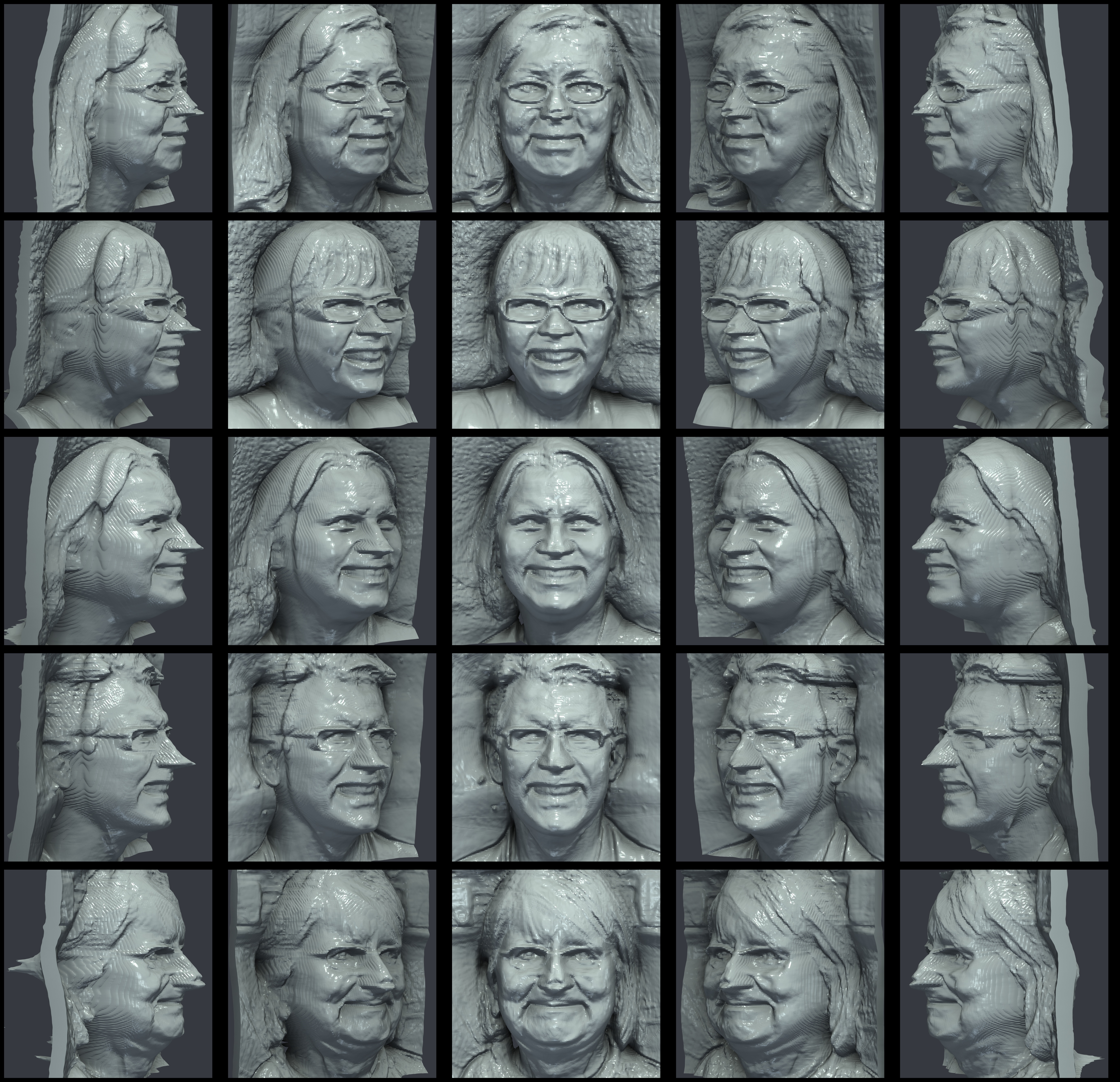}\\[1pt]{\small (d) point-cloud (PointNet) reward}
  \end{minipage}
  \caption{Geometry of the fine-tuned EG3D generator on the same seeds, at the same checkpoint, under four conditions: (a) no reward ($\lambda_r=0$, control), (b) the $\sigma_{XYZ}$ reward, (c) the triple-view depth-map reward, and (d) the point-cloud (PointNet) reward. Seeds $1$--$5$ (rows) are common to all runs. The $\sigma_{XYZ}$ reward reshapes geometry while preserving identity, whereas the no-reward control leaves geometry essentially unchanged and the weaker image-derived rewards (Table~\ref{tab:repaccuracy}) produce smaller or less coherent corrections.}
  \label{fig:geom_reward_compare}
\end{figure}

We fine-tune EG3D with our best reward model, the ResNet-SE-3D-UNet $\sigma_{XYZ}$ model (Table~\ref{tab:repaccuracy}). Fine-tuning results for this model are reported here; the less successful results from the other reward representations are shown in Figure~\ref{fig:geom_reward_compare}. Since our modified GAN loss now includes the reward loss $\mathcal{L}_{r}$, a control experiment is run to isolate the effect of the reward model on 3D geometry and 2D view fidelity as measured by FID. We fine-tune two versions of EG3D where either $\lambda_{r} = 0$ or $\lambda_{r} = 10$ in Equation~\eqref{eq:lg}, isolating the influence of the reward model with all other hyperparameter settings kept identical. As reported in Table~\ref{tab:fid}, the 2D image quality measured by FID-50k degrades in both experiments relative to the pretrained generator, but the larger degradation occurs when $\lambda_{r} = 10$. After fine-tuning for $20$ kimg ($\approx\!20{,}000$ images shown, at batch size $16$) the geometry of the $\lambda_{r} = 10$ experiment is clearly improved, and we use this checkpoint, $G_{r_{\theta}^{*}}$, for the external user study. We do not train beyond this point and make no claim of mode collapse: the truncation-baseline analysis of Section~\ref{sec:results:truncation} shows that the tuned geometry stays closer to each seed's original shape than to the truncation-mean face, so the reward selectively reshapes geometry rather than collapsing the distribution toward a common mean shape. This is consistent with the broader observation that reward-based fine-tuning trades a measure of output diversity for quality \citep{kirk2024rlhf}; in our setting the cost surfaces as a modest FID increase while 2D identity remains qualitatively similar (Figure~\ref{fig:before_after}) and the geometry is selectively corrected. At its final checkpoint, $G_{r_{\theta}^{*}}$ reaches an FID-50k of $6.657$, against $5.342$ for the matched $\lambda_{r} = 0$ control and $4.09$ for the untuned pretrained generator. Fine-tuning is therefore associated with an FID-50k increase of about $1.25$ (pretrained to control), while adding the reward loss is associated with a further increase of about $1.32$ (control to reward-tuned). On the other hand, with $\lambda_{r} = 0$ the 3D geometry does not change observably. Under the reward loss the reward improves steadily for essentially every latent code, whereas the control run shows no systematic reward change (Figure~\ref{fig:reward_traj}); the reward distribution after fine-tuning is shown in Figure~\ref{fig:reward_hist}.

\begin{figure}[htbp]
  \centering
  \includegraphics[width=\linewidth]{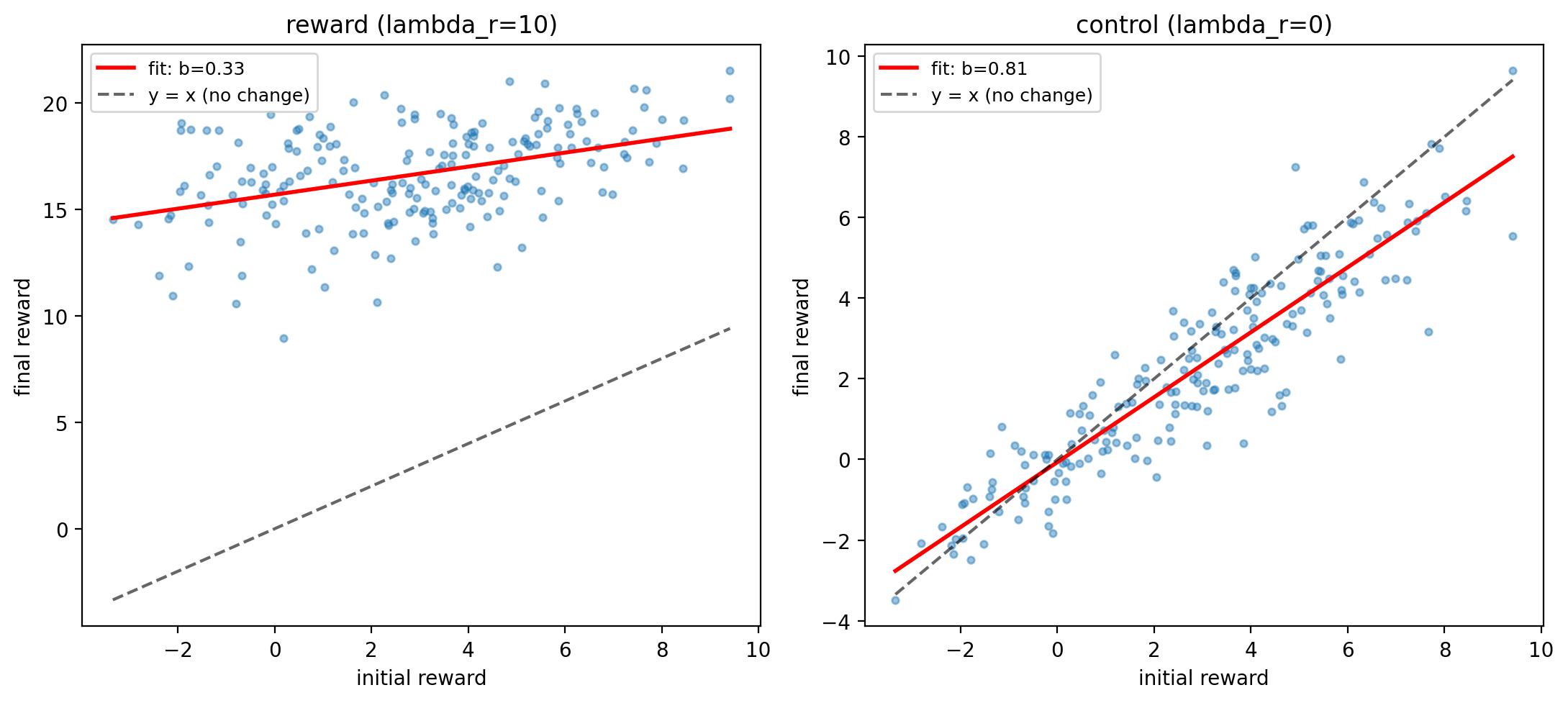}
  \caption{Final vs.\ initial $\sigma_{XYZ}$ reward for $200$ fixed latent codes; the dashed line is $y\!=\!x$ (no change). \emph{Left:} with the reward loss ($\lambda_{r}=10$) every code lies well above $y\!=\!x$ and the fit is nearly flat ($b = 0.33$), so the final reward is almost independent of the starting quality - the reward compresses the distribution toward a common high-quality level while improving all codes. \emph{Right:} the no-reward control ($\lambda_{r}=0$) stays on the identity line ($b = 0.81$). A slope $b < 1$ is the de-biased test for ``lower-quality samples improve more'', avoiding the regression-to-the-mean artefact of regressing the gain on the baseline.} \label{fig:reward_convergence}
\end{figure}

To quantify whether the reward gain depends on a sample's starting quality, we regress each seed's final reward on its initial reward across the $200$ codes (Figure~\ref{fig:reward_convergence}). Because the reward is deterministic for a fixed latent code, the appropriate de-biased test is whether this slope $b$ lies below $1$: $b < 1$ indicates that lower-quality samples improve more and the reward distribution compresses. Under the reward loss $b = 0.33$ ($95\%$ CI $[0.23, 0.43]$, $p \!\approx\! 3 \!\times\! 10^{-28}$ against $b = 1$) with a mean reward gain of $+13.8$: every code improves substantially and the final reward is nearly independent of the starting quality, so the reward signal pulls poor and good geometry alike toward a common high-quality level. The matched control stays near the identity line ($b = 0.81$, mean change $-0.6$), showing neither systematic improvement nor strong compression. We report the final-on-initial slope rather than regressing each seed's gain on its own baseline, as the latter conflates a real effect with a regression-to-the-mean artefact.

\begin{table}[t]
\caption{FID-50k (lower is better) of the pretrained EG3D generator and of the fine-tuned generators at the final checkpoint, with the reward loss ($\lambda_{r} = 10$) and the matched no-reward control ($\lambda_{r} = 0$), all evaluated against the same real-data statistics of the resynthesised FFHQ dataset. Fine-tuning is associated with an FID increase of $1.25$ (pretrained $\to$ control), and adding the reward loss is associated with a further $1.32$ increase (control $\to$ reward-tuned).}
\label{tab:fid}
\centering
\begin{tabular*}{\tblwidth}{@{}LL@{}}
\toprule
Configuration & FID-50k \\
\midrule
Pretrained EG3D (untuned)               & 4.092 \\
$\lambda_{r} = 0$ (no-reward control)  & 5.342 \\
$\lambda_{r} = 10$ (reward fine-tuning) & 6.657 \\
\bottomrule
\end{tabular*}
\end{table}

\subsubsection{External user study}\label{sec:results:user}

Improvements to 3D shape are evaluated via an external user study in which we present $n = 40$ face shapes before and after fine-tuning. We visualise pairs of 3D shapes: one from $G$ and the other from $G_{r_{\theta}^{*}}$, synthesised from a fixed latent code $z_{i}$, and ask $n = 17$ respondents whether they prefer either. Of the total $17 \!\times\! 40 = 680$ questions asked, $506$ responses showed a tuned preference, $141$ showed an original preference, and the remaining responses indicated neither. Results are analysed using Cohen's $h$, which compares two proportions in a multiple-choice questionnaire in order to indicate the degree of preference \citep{cohen2013statistical}. For our study, Cohen's $h \!=\! 1.135$, indicating a large effect size. The proportion of all responses is given in Table~\ref{tab:userstudy}. These results indicate that users overwhelmingly prefer 3D geometries after tuning with the reward model.

\begin{table}[t]
\caption{Summary of user-preference proportions in pairwise comparisons of 40 fine-tuned examples.}
\label{tab:userstudy}
\centering
\begin{tabular*}{\tblwidth}{@{}LL@{}}
\toprule
Outcome & Proportion \\
\midrule
$x_{G_{r_{\theta}}} \!\succ\! x_{G}$ & 0.744 \\
$x_{G} \!\succ\! x_{G_{r_{\theta}}}$ & 0.207 \\
No preference                       & 0.049 \\
\bottomrule
\end{tabular*}
\end{table}

\subsubsection{Before-and-after visualisations}\label{sec:results:beforeafter}

\begin{figure}[htbp]
  \centering
  \includegraphics[width=0.95\linewidth]{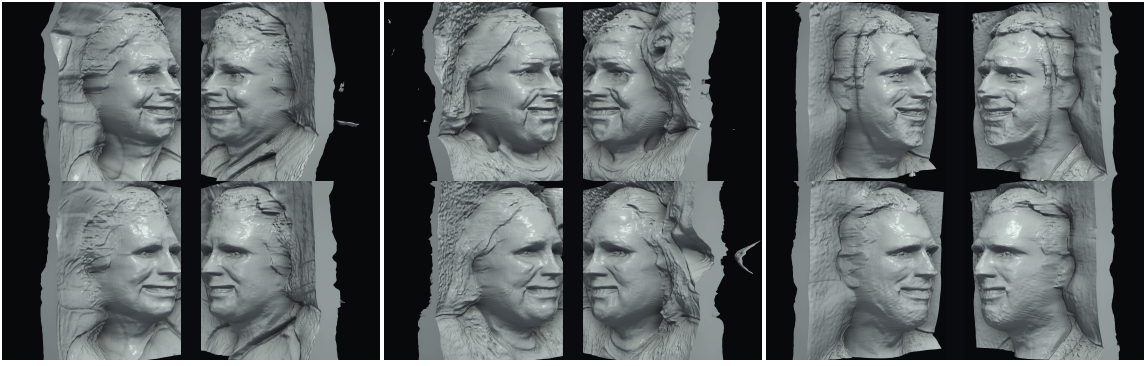}
  \caption{Change in geometry after fine-tuning for three fixed latent codes (seeds $200005$, $200025$, $200060$, arranged left to right). The upper row visualisations are sampled from the generator before fine-tuning. The lower row  visualisations are sampled after fine-tuning.}
  \label{fig:beforeafter}
\end{figure}

The changes in face geometry from the original $G$, compared to the fine-tuned $G_{r_{\theta}^{*}}$, are apparent via visual inspection in Figure~\ref{fig:beforeafter}. Examples drawn from $G$ which contain obvious defects, such as discontinuities on the nose or sides of the face, are observed to be improved after fine-tuning. Importantly, the general shape of the face is maintained. The increased FID score indicates a measurable distributional cost, but Figure~\ref{fig:before_after} shows that for a fixed latent code the rendered face remains qualitatively similar in identity and appearance between the start and end of fine-tuning, even as the underlying geometry is reshaped. Further mesh visualisations, including per-generator reward-ranked tails, are shown in Figure~\ref{fig:unstratified_mesh_tails}.

\subsection{Post-hoc analyses}\label{sec:results:posthoc}

We supplement the headline user study and qualitative comparisons with five further analyses that probe (i)~the learned structure of the depth-map and point-cloud reward backbones, (ii)~the structure of the learned $\sigma_{XYZ}$ reward embedding space, (iii)~the robustness of the reward improvement to identity matching, (iv)~whether fine-tuning is merely a collapse toward the truncation-mean face, and (v)~which face regions the $\sigma_{XYZ}$ reward model attends to. Throughout this subsection, $G$ denotes the pretrained EG3D generator and $G_{r_{\theta}^{*}}$ the generator after $20$ kimg of fine-tuning.\footnote{The numerical values reported in this subsection are computed on a representative fine-tuning run.}

\subsubsection{Robustness of reward improvement}\label{sec:results:robustness}

\paragraph{Same-latent comparison.}
For $100$ latent codes sampled with truncation $\psi = 0.7$, we compute the reward delta $\Delta s(z) = r_{\theta}(G_{r_{\theta}^{*}}(z)) - r_{\theta}(G(z))$. The empirical distribution has mean $+12.9$ and median $+12.6$, and the fraction of positive deltas is $1.0$ - the fine-tuned generator strictly dominates the pretrained generator in reward space on this latent sweep. The mean $L^{2}$ distance between paired $512$-d embeddings is $161$.

\paragraph{Identity-matched comparison.}
A more demanding test holds facial identity approximately constant across the two generators. We pre-compute reward-model embeddings for a bank of $5{,}000$ samples from the fine-tuned generator. For each of $500$ samples from the pretrained generator, we retrieve the nearest tuned-bank sample under cosine similarity in a face-recognition embedding, retaining only matches with cosine $\geq 0.80$. This yields $n = 24$ identity-matched pairs with mean identity cosine $0.84$. The reward delta on this matched set has mean $+11.9$, median $+11.6$, with all $24$ deltas positive. The mean $\sigma$-pair $L^{1}$ distance is $11.4$ despite the matched-identity constraint, and the mean latent-space distance between the matched $z$ codes is $31.5$ - the reward improvement is therefore not an artefact of identity drift, because it persists when identity is held fixed by an external face-recognition model.

\subsubsection{Comparison against the truncation baseline}\label{sec:results:truncation}

A natural alternative explanation for the reward gain is that fine-tuning simply pulls every sample toward the truncation mean face (which is, by construction, smoother and free of structural defects, with lower diversity). We test this by generating, for each of $100$ shared latent codes, three samples: the pretrained generator at $\psi = 0.7$ (denoted $x_{\text{orig}}$), the fine-tuned generator at $\psi = 0.7$ (denoted $x_{\text{tuned}}$), and the pretrained generator at $\psi = 0$ (denoted $x_{\text{trunc}}$). We then measure, in both depth-map and $\sigma_{XYZ}$ representations, whether $x_{\text{tuned}}$ is closer to $x_{\text{orig}}$ or to $x_{\text{trunc}}$.

\begin{table}[t]
\caption{Geometric comparison of the tuned generator against the pretrained generator at $\psi = 0.7$ and the truncation mean at $\psi = 0$. \emph{Closer to original} indicates the fraction of $n = 100$ shared latents for which the tuned sample is geometrically nearer to the original than to the truncated mean. Linear projection $\alpha$ is the scalar coefficient when projecting the tuned sample onto the $(x_{\text{trunc}} - x_{\text{orig}})$ direction; the residual fraction is the proportion of the tuning move that is orthogonal to this axis.}
\label{tab:truncation_baseline}
\centering
\begin{tabular*}{\tblwidth}{@{}LLLL@{}}
\toprule
Representation & Closer to original & Projection $\alpha$ & Residual fraction \\
\midrule
Depth map      & $0.98$ ($p \!\approx\! 2.6{\times}10^{-18}$) & $0.22$ & $0.88$ \\
$\sigma_{XYZ}$ & $0.93$ ($p \!\approx\! 1.4{\times}10^{-17}$) & $0.33$ & $0.92$ \\
\bottomrule
\end{tabular*}
\end{table}

As reported in Table~\ref{tab:truncation_baseline}, the tuned sample is closer to the original than to the truncation mean in $98\%$ of cases for the depth-map representation and $93\%$ for $\sigma_{XYZ}$, with one-sided Wilcoxon $p$-values below $10^{-17}$ in both cases. The linear projection of the tuning move $x_{\text{tuned}} - x_{\text{orig}}$ onto the truncation direction $x_{\text{trunc}} - x_{\text{orig}}$ has coefficient $0.22$ (depth) and $0.33$ ($\sigma$), and the residual fraction - the proportion of the tuning move that lies orthogonally to the truncation axis - is approximately $0.9$ in both representations. Fine-tuning is therefore not a collapse to the mean face: most of the geometric change is along a direction unavailable to the truncation operation.

\paragraph{Identity preservation.}
We also quantify perceptual and identity-level drift in RGB-space between $G$ and $G_{r_{\theta}^{*}}$. Mean LPIPS \citep{zhang2018lpips} between original and tuned views is $0.19$ on the canonical view and $0.20$ averaged over eight viewpoints. Identity cosine, measured by a pretrained face-recognition network, has mean $0.84$ on the canonical view and $0.82$ averaged over eight viewpoints (worst single viewpoint per pair: $0.72$). The within-model view-to-canonical identity consistency is $0.879$ for the pretrained generator and $0.869$ for the tuned generator, so the cross-model view-to-canonical consistency drops by $0.010$ cosine units relative to the pretrained baseline (Wilcoxon $p \!\approx\! 1.1 \!\times\! 10^{-5}$); the within-model pairwise view consistency falls from $0.820$ (pretrained) to $0.803$ (tuned), a drop of $0.017$ ($p \!\approx\! 1.5 \!\times\! 10^{-7}$). These drops are small but statistically resolvable, and represent an honest cost of the fine-tuning procedure. Crucially, Spearman correlations between identity drift and the magnitude of the geometric change are below $0.21$ in absolute value across all pairings tested, and only one of six reaches $p < 0.05$. The geometric improvement and identity drift are therefore approximately independent failure modes, not coupled ones.

\paragraph{Reward is truncation-aware, yet tuning is not mean-regression.}
A natural concern is that the reward model may simply have learnt to prefer the low-truncation ``mean'' face: the high-quality anchor $x_{HQ}$ added to every preference batch is itself a low-truncation sample ($\psi = 0.25$), drawn from the centre of the latent space and presented as the superlative example during training. (At $\psi = 0$ the generator collapses to a single mean face regardless of its noise input.) To test this, we sweep $\psi \in \{0.0, 0.25, 0.5, 0.7, 1.0\}$ on the same $100$ latent codes and score every regime with the $\sigma_{XYZ}$ reward (Table~\ref{tab:trunc_sweep}). The reward is strongly monotonic in truncation: $\bar{r} = +17.93$ at $\psi=0$ (the mean face) versus $\bar{r} = +2.78$ at $\psi=1.0$ (full diversity), a span of $\sim\!15$ reward units. Crucially, the tuned generator at $\psi = 0.7$ scores $\bar{r} = +18.64$ - equivalent to the trunc-$0$ mean face - yet the truncation-baseline analysis above shows that $98\%$ of tuned samples are geometrically closer to the original than to the mean face, with a projection of only $\alpha \approx 0.22\text{--}0.33$ along the mean-regression axis (residual fraction $\approx 0.9$). The reward model would in principle reward mean-regression, but the fine-tuning procedure finds geometric directions orthogonal to that axis that attain equivalent reward without collapsing identity. This is the strongest evidence against a trivial mean-regression interpretation of the reward gain.

\begin{table}[t]
  \centering
  \caption{$\sigma_{XYZ}$ reward score on the EG3D-orig generator across truncation $\psi$, on $100$ latent codes. The reward is monotonically decreasing in $\psi$; the tuned generator at $\psi=0.7$ scores $+18.64$, matching the trunc-$0$ mean face under the same reward but via geometric directions orthogonal to the mean-regression axis (Section~\ref{sec:results:image_reward} forwards this analysis to a cross-generator transfer setting).}
  \label{tab:trunc_sweep}
  \footnotesize
  \begin{tabular*}{\tblwidth}{@{}LLLL@{}}
    \toprule
    EG3D-orig $\psi$ & Mean reward & Median & Std \\
    \midrule
    $0.00$ (mean face)      & $+17.93$ & $+17.93$ & $\!\sim\!0$ \\
    $0.25$ (HQ regime)      & $+14.70$ & $+14.93$ & $1.50$ \\
    $0.50$                  & $\phantom{+}+8.70$ & $\phantom{+}+8.56$ & $2.36$ \\
    $0.70$ (canonical)      & $\phantom{+}+5.76$ & $\phantom{+}+5.80$ & $2.32$ \\
    $1.00$ (full diversity) & $\phantom{+}+2.78$ & $\phantom{+}+2.89$ & $2.40$ \\
    \midrule
    EG3D-tuned, $\psi=0.7$  & $+18.64$ & $+18.51$ & $1.90$ \\
    \bottomrule
  \end{tabular*}
\end{table}

\subsubsection{Analysis and interpretability of intermediate representations}\label{sec:results:shap}
We analyse the learned intermediate representations at two levels. The reward model's own learned $\sigma_{XYZ}$ embedding is examined to test whether it stratifies geometric quality more cleanly than the raw density feature. Second, because the single-view depth-map and point-cloud reward backbones perform comparatively poorly (Table~\ref{tab:repaccuracy}), their representations are inspected directly with SHAP-style attribution.

\paragraph{Reward-model embedding stratification.}
The learned representation can be characterised the two feature vectors the $\sigma_{XYZ}$ backbone produces per sample (Figure~\ref{fig:reward_arch}): the ResNet-SE-3D-UNet emits an $8{,}192$-dimensional global feature, which a multi-layer perceptron compresses to the $512$-dimensional vector $\bar{f}$ that the scoring heads read to produce the scalar reward. Drawing $100$ samples per regime under three truncation levels $\psi \!\in\! \{0.25, 0.70, 1.00\}$ of the pretrained generator, reward stratifies monotonically with $\psi$ ($22.1 \pm 0.9$, $18.6 \pm 1.9$ and $16.7 \pm 2.3$ respectively). The silhouette coefficient of the regime labels rises from $0.10$ in the raw $8{,}192$-d feature to $0.21$ in the compressed $512$-d $\bar{f}$. Quality is therefore more cleanly separated after compression than in the raw density feature, indicating the network has learned structure not given directly by the pairwise labels.

\begin{figure}[htbp]
  \centering
  \begin{subfigure}[t]{0.49\linewidth}
    \centering
    \includegraphics[width=\linewidth]{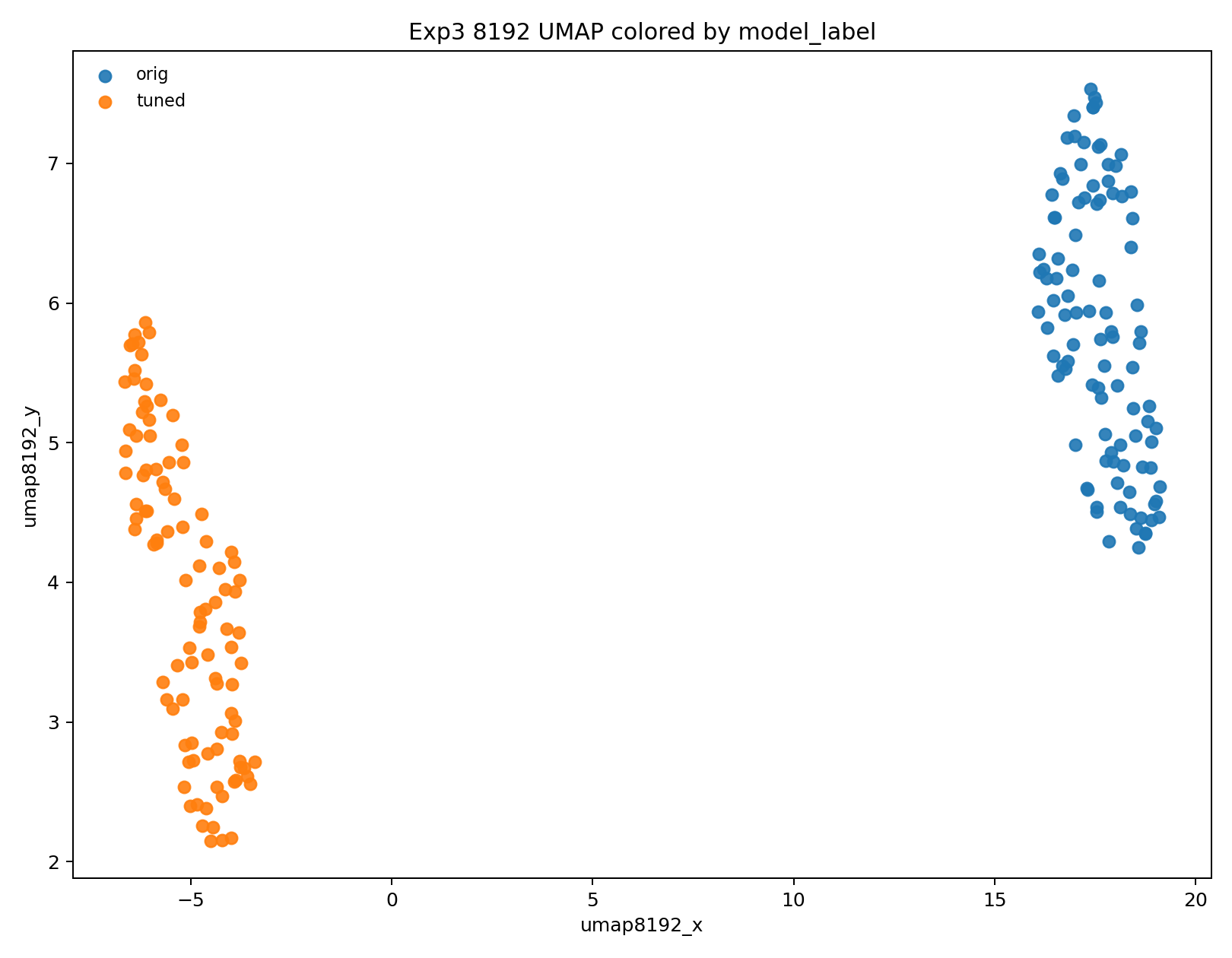}
    \caption{Coloured by model (orig vs tuned)}
    \label{fig:embed_umap}
  \end{subfigure}
  \hfill
  \begin{subfigure}[t]{0.49\linewidth}
    \centering
    \includegraphics[width=\linewidth]{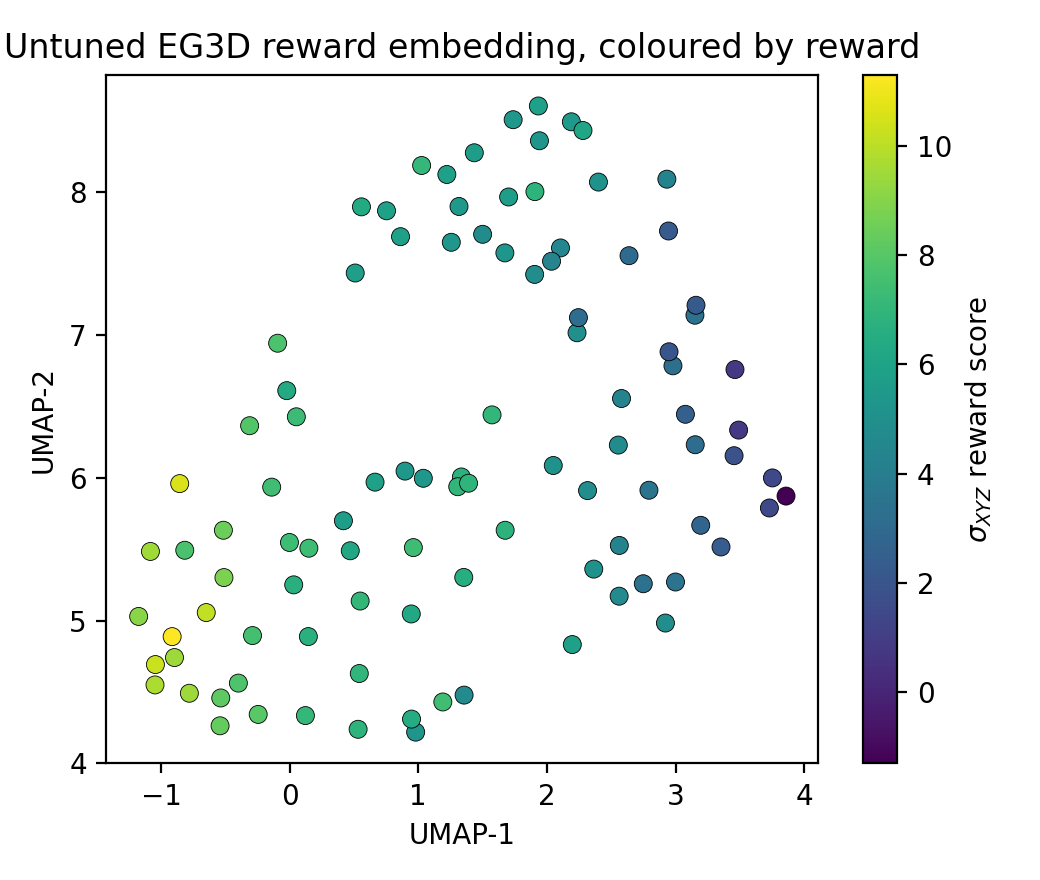}
    \caption{Untuned $G$, coloured by $\sigma_{XYZ}$ reward}
    \label{fig:embed_umap_reward}
  \end{subfigure}
  \caption{UMAP projections of the reward model's $8{,}192$-d global feature. \subref{fig:embed_umap}~Same-$z$ pairs from $G$ (orig) and $G_{r_{\theta}^{*}}$ (tuned), colour-coded by model; the two populations are cleanly separated under the learned feature. \subref{fig:embed_umap_reward}~$100$ samples from the untuned generator $G$, colour-coded by $\sigma_{XYZ}$ reward score; reward varies smoothly and monotonically across the embedding (Spearman $\rho = -0.93$ between reward and the principal UMAP axis).}
  \label{fig:embed_umap_both}
\end{figure}

Repeating the analysis on $100$ pairs $(G(z), G_{r_{\theta}^{*}}(z))$ sharing the same latent code $z$ (Section~\ref{sec:results:robustness}), the original and tuned populations separate with silhouette $0.49$ in the $8{,}192$-d feature and $0.80$ in the compressed $\bar{f}$. Figure~\ref{fig:embed_umap} visualises the $8{,}192$-d feature via UMAP, with the two populations cleanly separated. A complementary view on the untuned generator alone (Figure~\ref{fig:embed_umap_reward}) shows the $\sigma_{XYZ}$ reward varying smoothly and monotonically across this feature, suggesting it reflects continuous structure rather than acting purely as a binary original-versus-tuned classifier.

\paragraph{Depth-map and point-cloud attribution.}
Given the comparatively poorer performance of the single-view depth-map and point-cloud reward models, we perform further analysis via the SHAP framework \citep{lundberg2017shap} for model interpretability. Adopting a game-theoretic perspective, SHAP models separate image regions as contributing either a positive or negative contribution to the model output, based on Shapley values \citep{shapley1953}. This approach is explored to indicate regions of the 3D geometry which tend to increase or decrease the 3D quality score $s$: green regions are estimated to increase the score, red regions to decrease the score, and transparent regions to have no effect. Using high-quality samples $x_{HQ}$, our expectation is to observe green regions in areas which coincide with high-quality 3D facial features. This analysis uses the model-agnostic image explainer of \citet{lundberg2017shap}: for a single rendered sample, contiguous image regions are masked and replaced with a neutral reference fill - we evaluate both zero- and dataset-average replacement - and each region's marginal effect on the score is estimated by sampling over masking coalitions. It is therefore an absolute, single-sample attribution, and is distinct from the paired untuned-to-tuned region-swap coalition applied to the $\sigma_{XYZ}$ backbone in Section~\ref{sec:results:shapley_face}, whose per-region baseline is the untuned generator's own density rather than a zero or average fill.

\begin{figure}[htbp]
  \centering
  \begin{subfigure}[t]{0.32\linewidth}
    \centering
    \includegraphics[width=\linewidth]{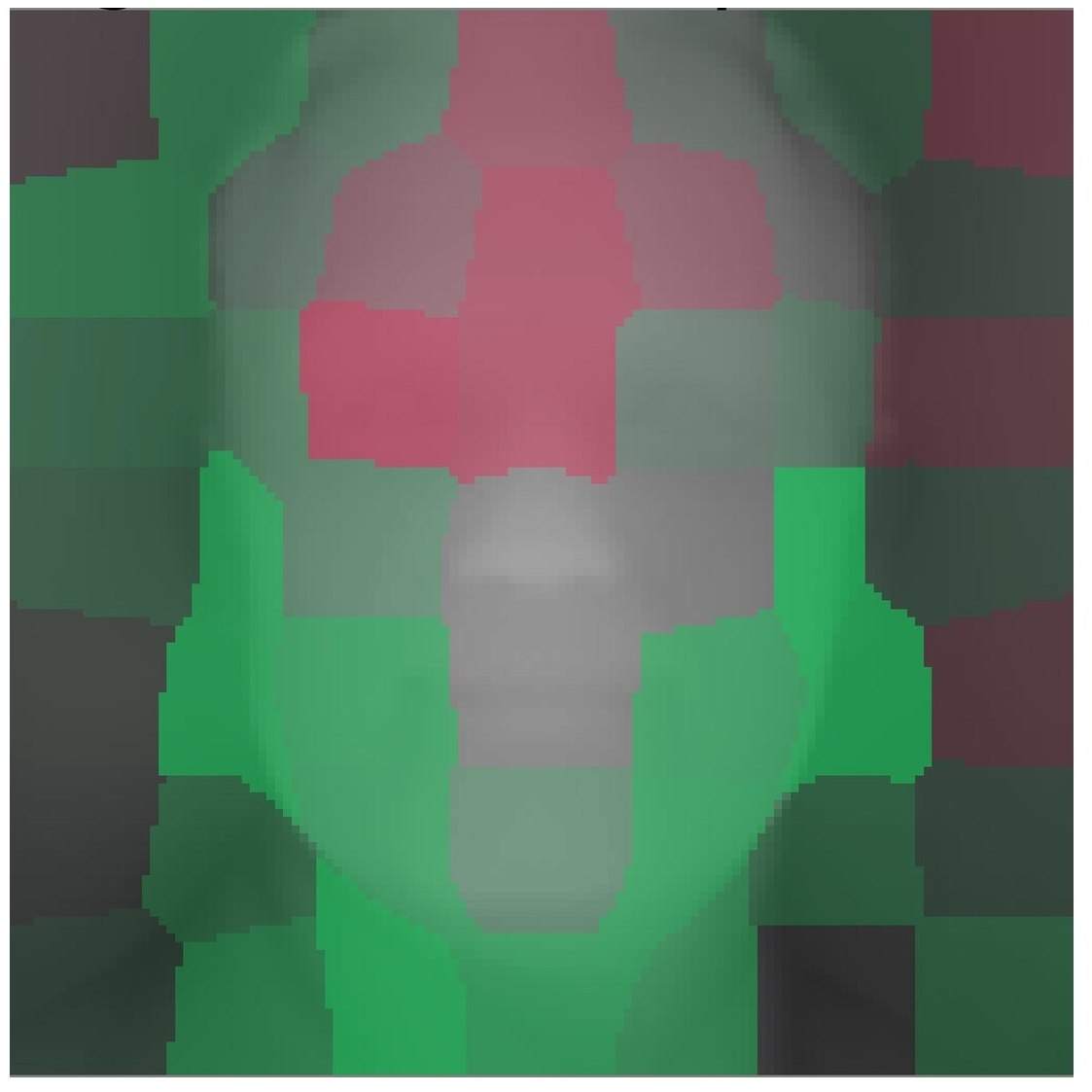}
    \caption{Depth map}
    \label{fig:shap_depth}
  \end{subfigure}
  \hfill
  \begin{subfigure}[t]{0.32\linewidth}
    \centering
    \includegraphics[width=\linewidth]{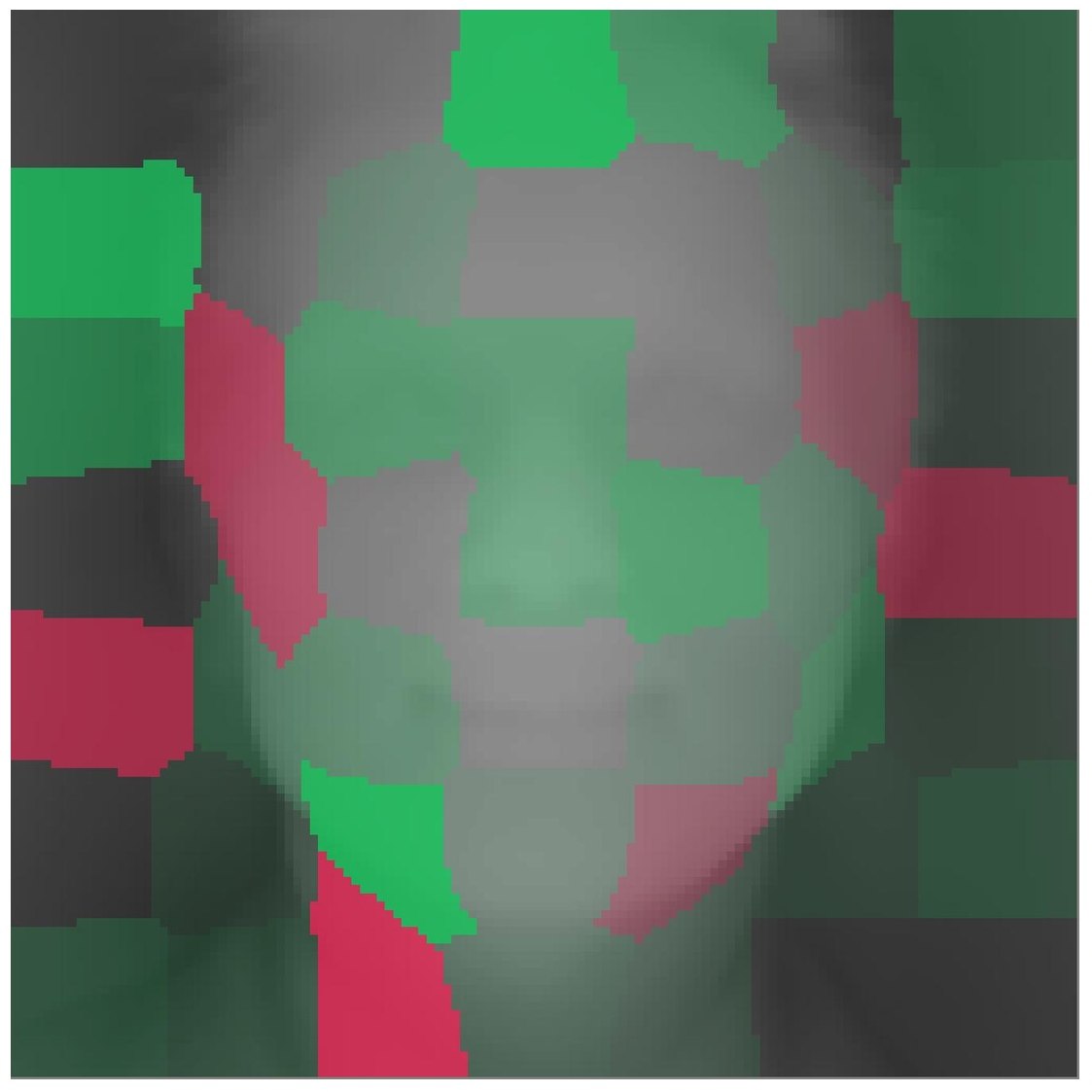}
    \caption{Point cloud}
    \label{fig:shap_pointcloud}
  \end{subfigure}
  \hfill
  \begin{subfigure}[t]{0.32\linewidth}
    \centering
    \includegraphics[width=\linewidth]{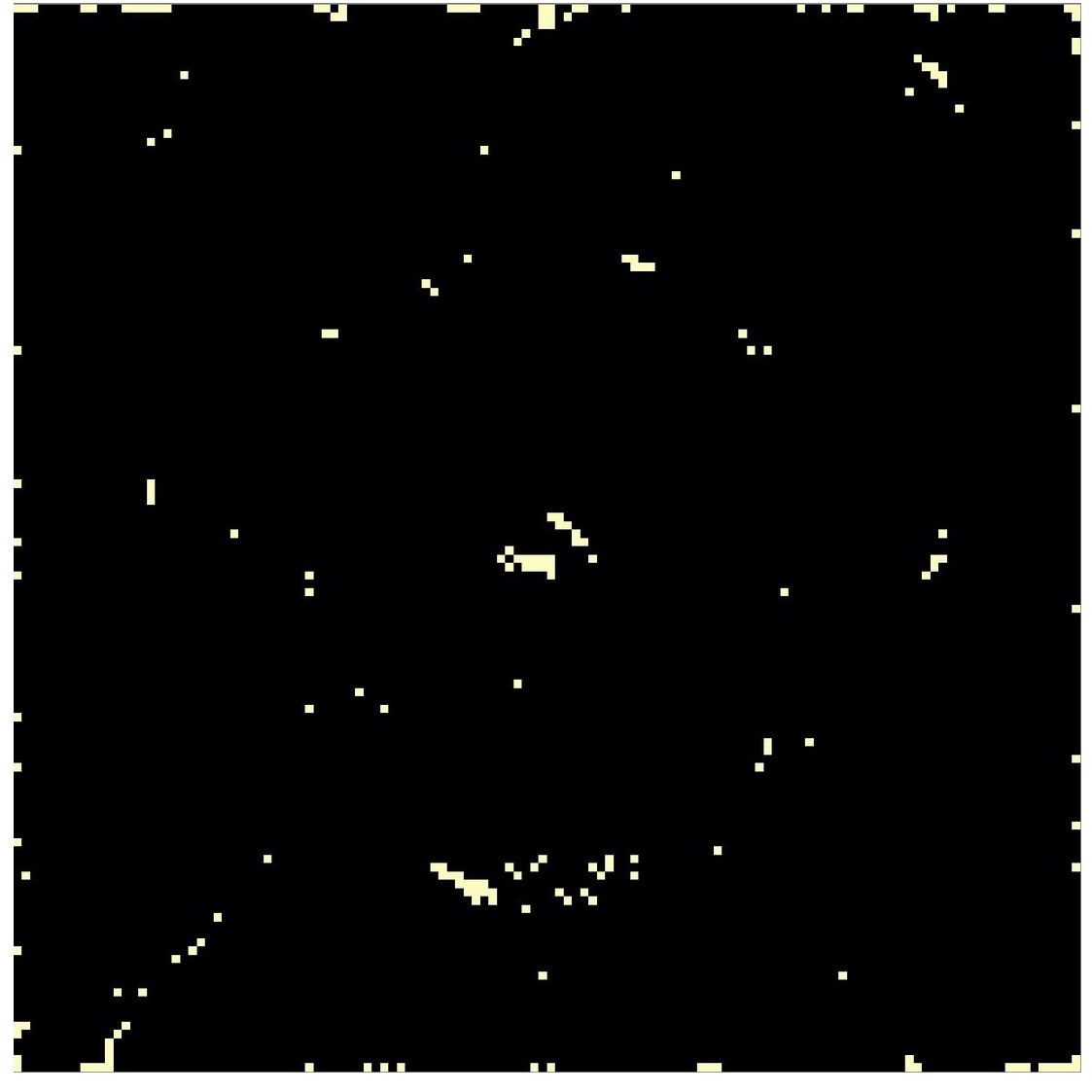}
    \caption{PointNet activations}
    \label{fig:shap_globalfeats}
  \end{subfigure}
  \caption{Visualisation of the contributions of facial regions to quality scores in depth-map and point-cloud reward models. \subref{fig:shap_depth}~SHAP-estimated contributions to quality score for the depth-map representation. \subref{fig:shap_pointcloud}~SHAP-estimated contributions to quality score for the point-cloud representation. \subref{fig:shap_globalfeats}~PointNet global max-activation points ($1024$ points as per the original PointNet architecture), which feed downstream reward losses, rarely overlap with facial regions. Both depth and point-cloud reward backbones attend to face edges away from semantic features; contrast with Figure~\ref{fig:region_shapley}, where the $\sigma_{XYZ}$ backbone attends to nose, mouth and cheeks.}
  \label{fig:shap}
\end{figure}

\paragraph{Depth map.}
As depicted in Figure~\ref{fig:shap}~(a), depth-map reward models tend to associate regions around the side of the face and very edge of the depth map with a higher reward score. The neutral or negative contributions of high-quality geometric regions, such as the nose tip or forehead, suggest that the model is not focusing on desired features.

\begin{figure}[htbp]
  \centering
  \begin{subfigure}{\linewidth}
    \centering
    \includegraphics[width=0.68\linewidth]{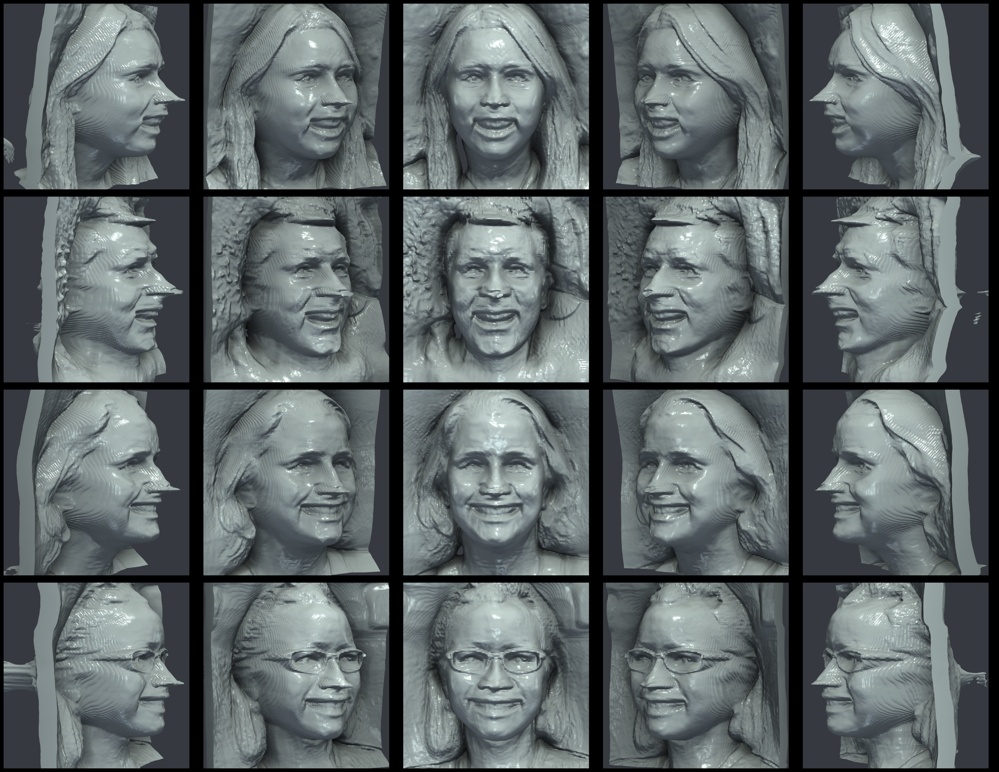}
    \caption{Highest-ranked (preferred) geometries}
    \label{fig:pointnet_top}
  \end{subfigure}
  \\[10pt]
  \begin{subfigure}{\linewidth}
    \centering
    \includegraphics[width=0.68\linewidth]{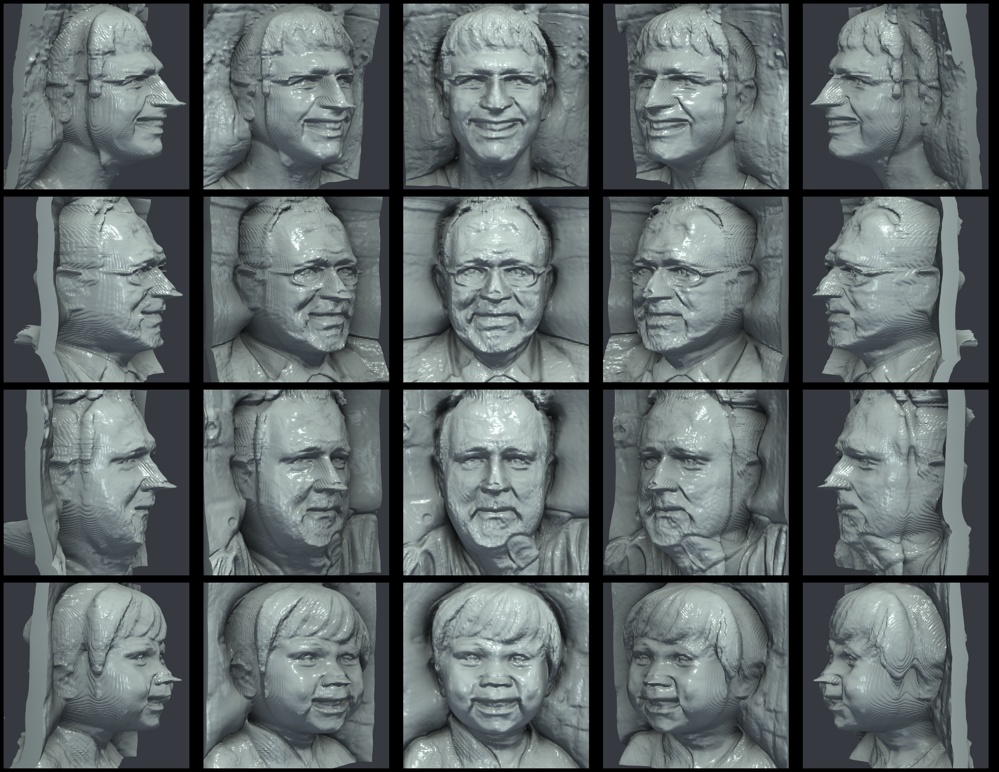}
    \caption{Lowest-ranked geometries}
    \label{fig:pointnet_bottom}
  \end{subfigure}
  \caption{Geometries ranked highest \subref{fig:pointnet_top} and lowest \subref{fig:pointnet_bottom} by the PointNet point-cloud reward, sampled from the EG3D generator after PointNet-reward fine-tuning. The top-ranked - i.e.\ preferred - geometries still include clearly defective shapes such as over-sharp noses and irregular surfaces, consistent with the model's near-chance within-distribution accuracy (Table~\ref{tab:repaccuracy}) and its edge-focused attribution (Figure~\ref{fig:shap}): the PointNet reward orders samples only weakly by genuine 3D quality.}
  \label{fig:pointnet_topbottom}
\end{figure}

\paragraph{Point cloud.}
Conversion of depth-map images to point clouds via Equation~\eqref{eq:pointcloud} permits us to conduct SHAP analyses for reward models using the PointNet backbone. As depicted in Figure~\ref{fig:shap}~(b), point clouds exhibit similar issues to depth maps, whereby regions on face sides contribute strongly to positive reward scores. Further insight can be developed by examining a heatmap of the $1024$ points which contribute to the global feature vector of the model, since all other points are excluded and hence ignored in appraisals of 3D shape quality. As depicted in Figure~\ref{fig:shap}~(c), the spatial locations of the activating points (highlighted in white) frequently occur around the square edge of the point cloud. Important regions on the face are discarded, implying that the model's evaluation of 3D shape quality is not concentrated on desired features. Figure~\ref{fig:pointnet_topbottom} shows the downstream consequence for fine-tuning: the geometries this reward ranks highest still include clearly defective shapes.

These weak image-derived rewards also tend to exaggerate geometric defects rather than correct them when used as a fine-tuning signal. For the point-cloud and single-view depth-map rewards we found that, if the reward is allowed to push the generator without the stabilising reward loss clamp introduced above, runaway reward values produce pronounced distortions - most visibly the nose being pulled progressively toward the camera as further reward updates are applied to the EG3D weights - yielding geometries that lie well off the distribution of meshes on which the reward model was trained. Because these rewards respond to face-edge regions rather than genuine quality cues (Figures~\ref{fig:shap} and~\ref{fig:pointnet_topbottom}), maximising them tends to drive the generator into out-of-distribution geometries that score highly yet are clearly degraded; the reward clamp was added in part to suppress this failure mode.

\subsubsection{Sigma-field reward attribution by face region}\label{sec:results:shapley_face}
Section~\ref{sec:results:shap} examined the depth-map and point-cloud reward backbones via SHAP and found them to attend to face edges away from features. We now apply Shapley values and Integrated Gradients \citep{lundberg2017shap,shapley1953,sundararajan2017ig} directly to the $\sigma_{XYZ}$ reward backbone on $100$ same-latent pairs and partition the canonical-view $\sigma$-volume slab into thirteen anatomically grounded regions. The attribution is paired rather than absolute: for each latent code the untuned generator's $\sigma$-cube is the reference and the tuned generator's $\sigma$-cube the target, and the resulting reward change is attributed across regions. For the Shapley estimate, regions are added to a coalition in random permutations by replacing their voxels with the tuned-generator values one region at a time, and each region's marginal reward change is averaged over permutations; by construction the per-region contributions sum to the full untuned-to-tuned reward delta. Integrated Gradients integrates the reward gradient along the straight-line path between the untuned and tuned cubes. Each region's baseline is therefore the untuned generator's own density in that region, not a zero, noise, or distribution-mean fill. Region masks are built by running the WFLW $98$-point landmark detector \citep{wu2018wflw,jin2021pipnet} on each seed's canonical-view render of the untuned generator, back-projecting every landmark to world coordinates via the EG3D canonical-view cam2world and pinhole intrinsics, averaging across the $100$ seeds, and forming axis-aligned bounding boxes from the WFLW semantic groupings (jawline, brows, eyes, nose bridge and bottom, outer and inner mouth) with a margin extending into the head interior. A forehead region is extrapolated above the brow landmarks using the brow-to-nose-tip proportion.

Two additional diagnostic non-landmark regions are defined: a front-of-camera band (voxels forward of the front-most landmark, restricted to the face $(x, y)$ rectangle) intended to detect the failure mode in which the reward gradient grows density into empty space, and a background-rear band. The named anatomical regions cover $42.7\%$ of the cube but absorb $88.6\%$ of the mean Shapley contribution, leaving only $8.6\%$ in a residual ``other'' bucket - a substantial improvement over the original $9$-region axis-aligned bounding boxes, which left $29\%$ of the reward delta in the residual.

\begin{figure}[htbp]
  \centering
  \includegraphics[width=\linewidth]{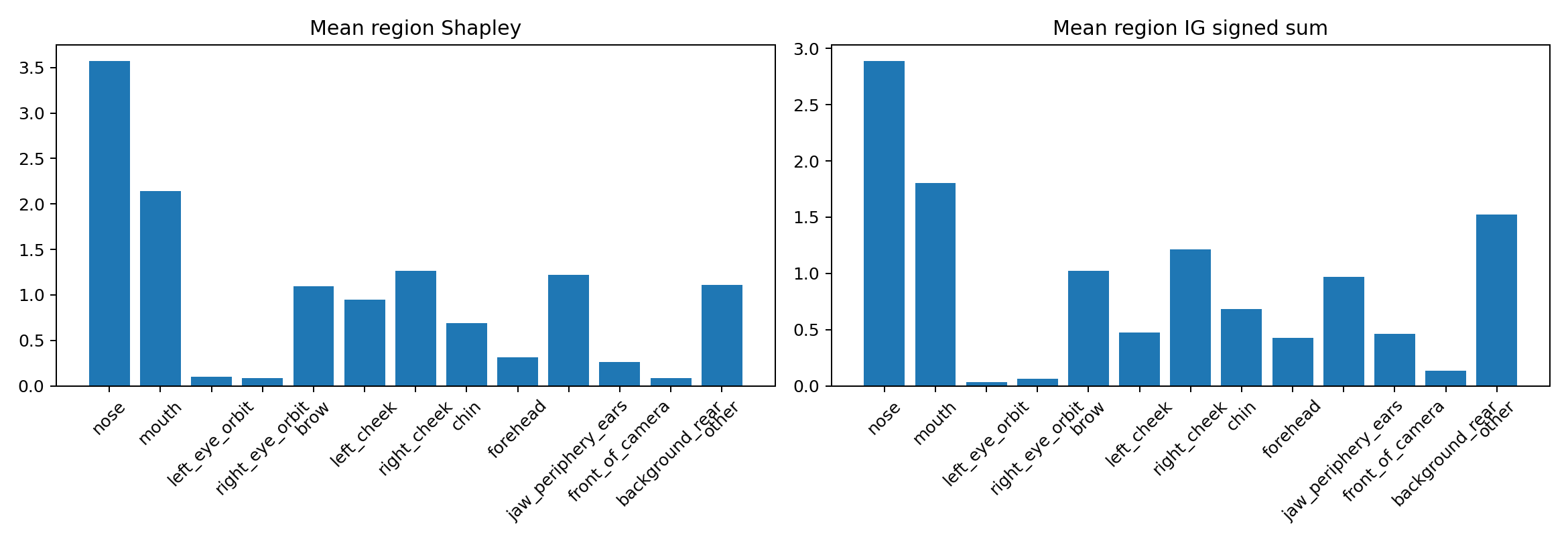}
  \caption{Mean Shapley (left) and Integrated Gradients (right) contribution per region for the $\sigma_{XYZ}$ reward model on $100$ identity-paired before/after seeds. Regions are derived from $98$-point WFLW landmarks \citep{wu2018wflw,jin2021pipnet} averaged across the seeds and back-projected to the $\sigma$-cube; the rightmost two bars (front\_of\_camera, background\_rear) are diagnostic non-landmark bands. Anatomically named regions account for $88.6\%$ of the mean reward delta.}
  \label{fig:region_shapley}
\end{figure}

The attribution (Figure~\ref{fig:region_shapley}) is dominated by the nose (mean Shapley $3.57$; top-$1$ region in $82/100$ seeds), followed by mouth ($2.14$; top-$1$ in $16/100$), right and left cheek ($1.26$, $0.95$), jaw periphery and ears ($1.22$), brow ($1.10$), chin ($0.69$), and forehead ($0.31$). Eye orbits contribute marginally (left $0.10$, right $0.08$; top-$1$ in $0/100$ seeds for both). Integrated Gradients yields the same ordering. Across the $100$ identity-paired seeds, every named region except brow has positive mean Shapley in at least $95\%$ of seeds.

The diagnostic front-of-camera region carries a small but non-trivial mean Shapley of $0.26$ (mean IG $0.46$), corresponding to roughly $2$--$3\%$ of the total reward delta on average, with peak per-seed values of $0.64$ Shapley and $1.11$ Integrated Gradients. On a subset of seeds the reward gradient places a small amount of density forward of the actual face surface. Under the EG3D legacy marching-cubes extraction (level $10$) used for our meshes this does not surface as a visible floating artefact in the rendered geometry, and the contribution is small relative to the named facial regions; we report it here only as a minor diagnostic.

This result is the direct counterpart of the SHAP analysis on the depth-map and point-cloud backbones in Section~\ref{sec:results:shap}, and provides insights regarding how the reward model computes a quality score. For $\sigma_{XYZ}$ the reward model attends predominantly to the nose, mouth, and cheeks, whereas when expressed on depth maps or point clouds it attends to face edges away from these features. The negligible contribution of the eye-orbit regions (top-$1$ in $0/100$ seeds) indicates that the reward signal does not strongly discriminate between alternative eye geometries in this sample, in contrast to the nose, mouth and cheek regions where the tuned-versus-untuned geometric change is concentrated.

\subsection{Generalisation of the reward}\label{sec:results:generalise}
A reward trained on EG3D's $\sigma$ field raises the question of how far it carries to other reward models and other generators. We probe this from three angles: agreement with pretrained image-based reward models, transfer across EG3D-family architectures, and a reward-guided inversion test. These analyses are diagnostic - they characterise the reward's scope rather than add to the fine-tuning result of Section~\ref{sec:results:finetune}, and the cross-architecture conclusions should be read as holding under the current reward and crop convention. Concretely, that convention means the same EG3D-derived frontal slab $X[64{:}192], Y[64{:}205], Z[102{:}231]$ and the same per-cube min-max rescaling to $[0,100]$ described in Section~\ref{sec:method:features} are reused before reward scoring.

\subsubsection{Agreement with image-based reward models}\label{sec:results:image_reward}

To position the $\sigma_{XYZ}$ reward signal against the contemporary wave of image-based 3D reward models, we score the same $100$ identity-paired seeds using two pretrained external reward models that operate on rendered multi-view imagery rather than on the density field directly. \textbf{MVReward} \citep{wang2025mvreward} consumes a canonical reference image and a bank of off-canonical views, returning a scalar multi-view reward; it requires no text prompt at inference. \textbf{Reward3D} \citep{ye2024dreamreward}, the reward backbone of DreamReward, consumes a four-view bank conditioned on a text prompt; the prompt is held fixed at a canonical face description across the original and tuned generators. This does not eliminate prompt sensitivity, but it isolates the before/after generator delta under one prompt choice. For each seed, both generators are rendered at the camera bank expected by the corresponding reward model, scored, and the per-seed reward delta $r^{\text{img}}_{\text{tuned}}(z) - r^{\text{img}}_{\text{orig}}(z)$ is recorded.

\begin{table}[t]
  \centering
  \caption{Reward delta statistics and pairwise Spearman correlations across $100$ identity-paired seeds for three reward models on the same EG3D before/after pair. $\sigma_{XYZ}$ scores all $100$ tuned generators as improved over the baseline; Reward3D (DreamReward) agrees in $77/100$ seeds; MVReward disagrees in $61/100$ seeds.}
  \label{tab:image_reward}
  \footnotesize
  \begin{tabular}{lcccc}
    \toprule
    Reward & mean $\Delta r$ & frac.\ positive & std & Spearman vs $\sigma_{XYZ}$ \\
    \midrule
    $\sigma_{XYZ}$ (ours)         & $+12.89$ & $1.00$ & $2.38$ & $-$ \\
    Reward3D \citep{ye2024dreamreward}    & $+0.10$ & $0.77$ & $0.18$ & $+0.25$ \\
    MVReward \citep{wang2025mvreward}     & $-0.03$ & $0.39$ & $0.09$ & $-0.05$ \\
    \bottomrule
  \end{tabular}
\end{table}

Two findings emerge from this diagnostic (Table~\ref{tab:image_reward}). First, under the fixed prompt used here, Reward3D agrees that the tuned generator is improved over the baseline in a majority of seeds ($77\%$), with a weak but positive Spearman correlation of $+0.25$ against the $\sigma_{XYZ}$ delta. Reward3D's own response distribution is within its training range on face renders ($\bar{r}_{\text{orig}}=0.60$, $\bar{r}_{\text{tuned}}=0.70$), so the agreement is not an artefact of saturation. Second, MVReward returns a near-zero mean reward delta and is essentially uncorrelated with $\sigma_{XYZ}$ ($\rho=-0.05$); both the orig and tuned generators score in the negative-reward region for MVReward ($\bar{r}_{\text{orig}}=-0.33$, $\bar{r}_{\text{tuned}}=-0.36$), consistent with the model being out-of-distribution on FFHQ-domain face renders. When instead ranking samples within a single generator, the two image-based rewards are moderately correlated with each other (Spearman $\approx 0.5$ on the EG3D-orig seed bank), consistent with both responding to the same rendered 2D appearance; what neither tracks is the $\sigma_{XYZ}$ reward, which operates in 3D space directly. This suggests that the density-volumetric framing retains a distinct role in the preference-tuning landscape and is not captured by these image-based alternatives under the present evaluation setup.

\subsubsection{Cross-generator transfer of the $\sigma_{XYZ}$ reward}\label{sec:results:crossgen}

\begin{figure}[htbp]
  \centering
  \includegraphics[width=\linewidth]{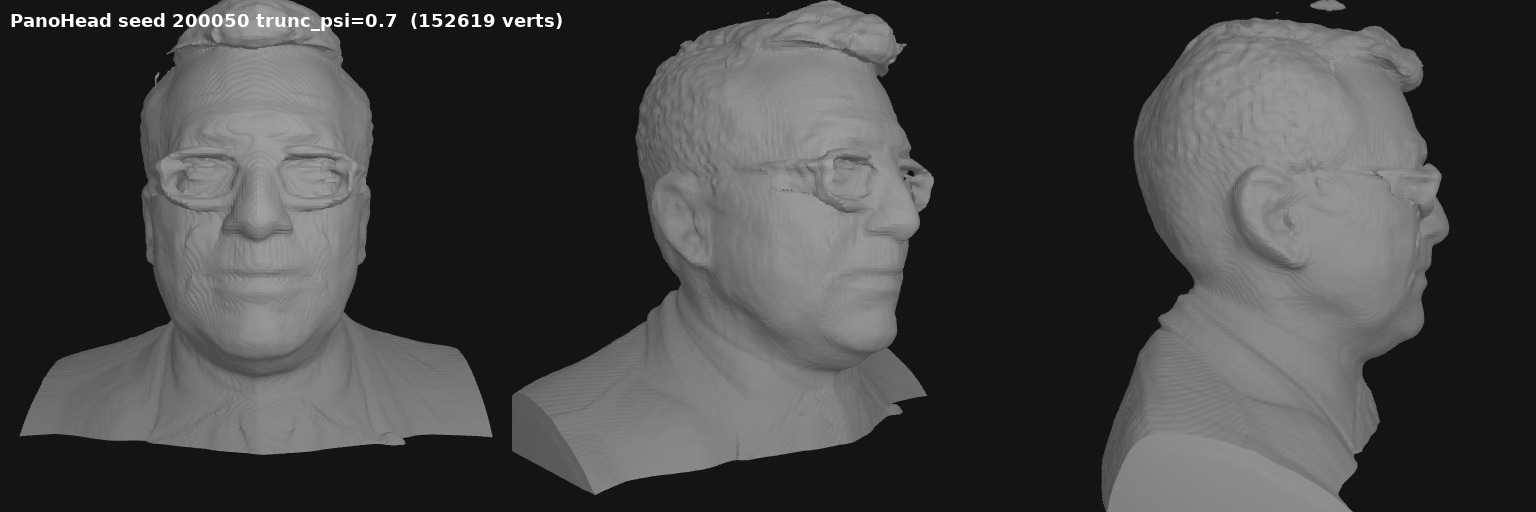}
  \caption{Marching-cubes mesh from a representative PanoHead sample (seed $200050$, $\psi=0.7$). Front, $45^{\circ}$ and $90^{\circ}$ views. PanoHead's tri-grid generator produces a coherent $360^{\circ}$ head with hair, ears, neck, shoulders and accessories. The visible mesh quality is not the source of the reward disagreement reported in Table~\ref{tab:trunc_sweep_panohead}; the disagreement reflects the different numerical scale of the two generators' $\sigma$ fields (Figure~\ref{fig:sigma_histogram}).}
  \label{fig:panohead_mesh}
\end{figure}

A complementary question is whether the $\sigma_{XYZ}$ reward signal transfers to a structurally different volumetric face generator. We score $100$ same-seed samples from PanoHead \citep{an2023panohead}, a tri-grid $360^{\circ}$ full-head generator trained on FFHQ-F, under the same reward + crop convention that was used for EG3D-orig (Table~\ref{tab:trunc_sweep_panohead}). PanoHead's RGB renders and marching-cubes are visually coherent $360^{\circ}$ heads with hair, ears, neck and accessories (Figure~\ref{fig:panohead_mesh}).

The truncation response on PanoHead is directionally ordered: moving from $\psi = 0.7$ to $\psi = 0$ raises the mean reward monotonically from $-5.40$ to $-4.85$. That direction matches EG3D, in line with the preference labels having favoured low-truncation HQ samples (Section~\ref{sec:method:dataset}). More importantly, it is not unique to PanoHead: the broader within-generator rank analysis later in Figure~\ref{fig:within_generator_rank_spearman} and Table~\ref{tab:within_generator_rank} shows that all tested generators retain a positive within-distribution rank signal, albeit with substantially different strength.

\begin{figure}[htbp]
  \centering
  \includegraphics[width=0.92\linewidth]{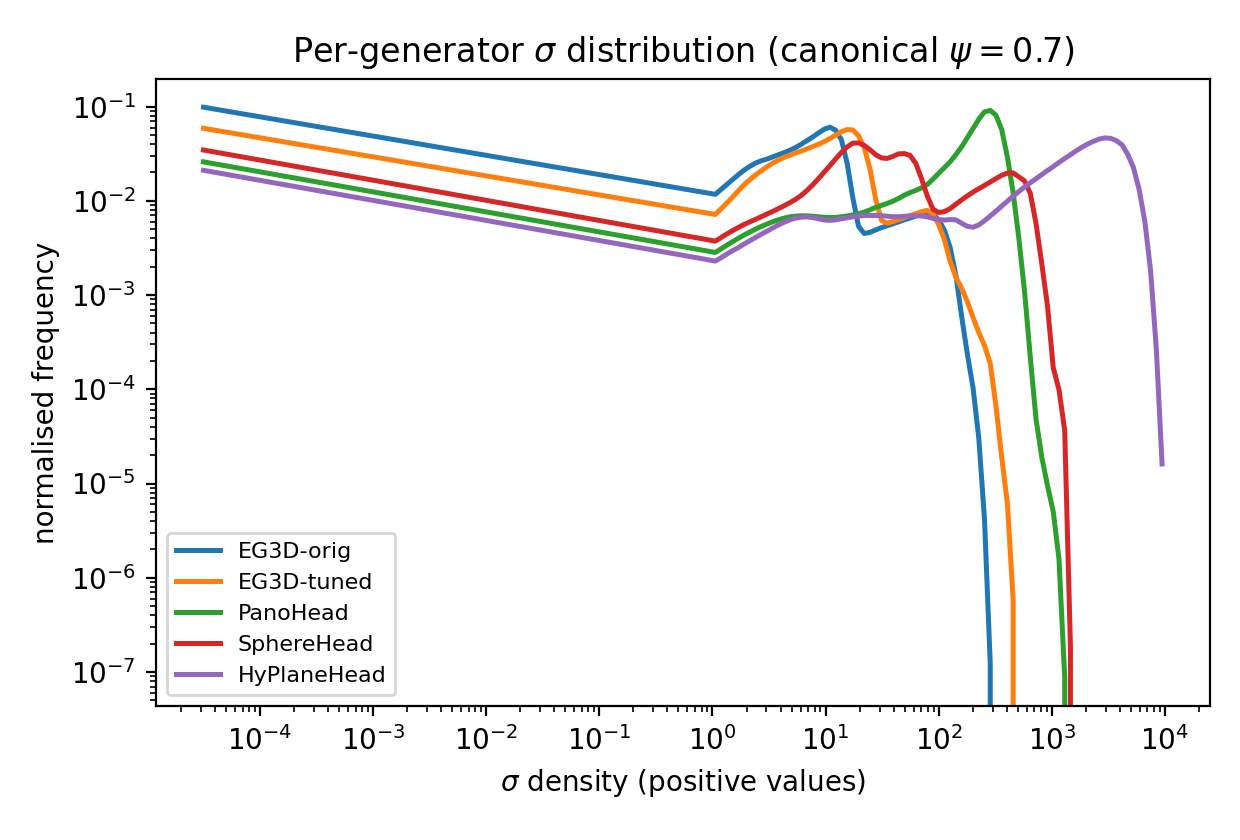}
  \caption{Distribution of positive $\sigma$ density across the five generators (canonical $\psi=0.7$, log--log axes). The geometry-bearing high-$\sigma$ tail occupies a markedly different numerical range per architecture: mean per-seed maximum $\sigma$ of roughly $250$ (EG3D-orig), $430$ (EG3D-tuned), $810$ (PanoHead and SphereHead) and $8{,}200$ (HyPlaneHead). A reward model trained on EG3D-FFHQ's $\sigma$ statistics is consequently out-of-distribution on the $360^{\circ}$ generators, which accounts for its compressed reward range there.}
  \label{fig:sigma_histogram}
\end{figure}

However, two quantitative differences are striking. First, the dynamic range of the reward on PanoHead is compressed by a factor of $\sim\!22$: the same $\psi$ sweep produces a span of only $\Delta\bar{r} \approx +0.55$ on PanoHead versus $\Delta\bar{r} \approx +12.17$ on EG3D-orig. The absolute reward magnitudes are separated by $\sim\!11$ reward units, with no overlap between the two $100$-seed distributions (PanoHead $\max = -2.6$ versus EG3D-orig $\min = -1.3$) - according to the reward scores, all geometries from EG3D-orig are worse than those from Panohead. It seems that this is more consistent with PanoHead's $\sigma$ field occupying a different numerical regime from EG3D-FFHQ's. Figure~\ref{fig:sigma_histogram} compares the raw $\sigma$ distributions of all five generators: the mean per-seed maximum density differs by more than an order of magnitude across architectures - roughly $250$ for EG3D-orig, $810$ for PanoHead and SphereHead, and $8{,}200$ for HyPlaneHead - so the high-$\sigma$ tail that the reward keys on is sharply displaced. A reward trained on EG3D's $\sigma$ statistics is therefore evaluated out-of-distribution on these generators. The per-sample $\mathtt{normalise\_sigma\_self}$ augmentation, which rescales each sigma cube to a common internal range before scoring, does not remove the mismatch.

Because the fine-tuning reward loss is clamped to $[-10, +10]$ (Section~\ref{sec:method:finetune}), the absolute reward magnitude does not by itself preclude using the $\sigma_{XYZ}$ reward to fine-tune PanoHead - the gradient is bounded regardless of where the initial reward distribution sits. The $22\times$ compressed dynamic range is the substantive limitation: under the current reward and crop convention, the reward provides correspondingly less signal about geometric quality on PanoHead's representation. A reward trained directly on PanoHead-domain preferences may be more performant. The broader finding of this section is that our best-performing 3D reward model is bound to the EG3D-sampled sigma field distribution on which it was trained.

\begin{table}[t]
  \centering
  \caption{$\sigma_{XYZ}$ reward score on PanoHead's $360^{\circ}$ full-head generator across truncation $\psi$, on the same $100$ latent codes used elsewhere. The low-$\psi$ to high-$\psi$ ordering matches the EG3D pattern of Table~\ref{tab:trunc_sweep}, but this table alone does not establish the full within-generator rank behaviour; that broader all-generator comparison is given in Figure~\ref{fig:within_generator_rank_spearman} and Table~\ref{tab:within_generator_rank}. The main point here is that PanoHead's reward dynamic range across the $\psi$ sweep is $22\times$ compressed compared to EG3D-orig, and the absolute magnitudes are separated by $\sim\!11$ reward units with zero overlap between the two distributions.}
  \label{tab:trunc_sweep_panohead}
  \footnotesize
  \begin{tabular*}{\tblwidth}{@{}LLLL@{}}
    \toprule
    PanoHead $\psi$ & Mean reward & Median & Std \\
    \midrule
    $0.00$ (mean face)         & $-4.85$ & $-4.85$ & $\!\sim\!0$ \\
    $0.25$ (HQ-regime analogue) & $-4.88$ & $-4.79$ & $0.36$ \\
    $0.70$ (canonical)         & $-5.40$ & $-5.58$ & $1.12$ \\
    \midrule
    $\psi$-sweep span ($0.7 \to 0.0$) & $+0.55$ & - & - \\
    (EG3D-orig equivalent span) & $+12.17$ & - & - \\
    \bottomrule
  \end{tabular*}
\end{table}

\paragraph{Canonical-truncation reward across four generators.}
The PanoHead comparison generalises to the wider EG3D family. We score $100$ same-seed samples at the canonical truncation $\psi = 0.7$ from two further $360^{\circ}$ architectures - SphereHead \citep{li2024spherehead} and HyPlaneHead \citep{li2026hyplanehead} - under the identical reward and crop convention used for EG3D-orig and PanoHead, and compare all four generators against the EG3D-orig baseline (Table~\ref{tab:cross_gen_canonical}). Every non-EG3D generator receives markedly lower $\sigma_{XYZ}$ reward: HyPlaneHead $-1.33$, SphereHead $-3.13$ and PanoHead $-5.40$, against EG3D-orig $+5.76$. Expressed in units of the EG3D-orig spread ($\mathrm{std} = 2.32$), the three architectures sit $3.1\sigma$, $3.8\sigma$ and $4.8\sigma$ below the baseline respectively. The separation is near-total: for PanoHead the two $100$-seed distributions do not overlap at all ($\max = -2.64$ falls below the EG3D-orig $\min = -1.31$), while for HyPlaneHead and SphereHead only their extreme upper tails ($\max = -0.24$ and $-0.99$) reach into EG3D-orig's lower tail. Under the current reward and crop convention, this ordering is consistent with the reward being bound to the representation on which it was trained rather than to an architecture-agnostic notion of geometric quality. 

\begin{table}[t]
  \centering
  \caption{$\sigma_{XYZ}$ reward at canonical truncation $\psi=0.7$ across four EG3D-family generators, on $100$ same-seed samples each under an identical reward and crop convention. Every $360^{\circ}$ architecture scores well below the EG3D-orig baseline; the final column reports the gap to the baseline in units of the EG3D-orig standard deviation ($2.32$). Reward model \texttt{7wnzkgie}.}
  \label{tab:cross_gen_canonical}
  \footnotesize
  \begin{tabular*}{\tblwidth}{@{}LLLLL@{}}
    \toprule
    Generator & Mean & Median & Std & Gap (EG3D $\sigma$) \\
    \midrule
    EG3D-orig    & $\phantom{+}+5.76$ & $\phantom{+}+5.80$ & $2.32$ & - \\
    HyPlaneHead  & $-1.33$ & $-1.40$ & $0.38$ & $3.1\sigma$ \\
    SphereHead   & $-3.13$ & $-3.08$ & $0.55$ & $3.8\sigma$ \\
    PanoHead     & $-5.40$ & $-5.58$ & $1.12$ & $4.8\sigma$ \\
    \bottomrule
  \end{tabular*}
\end{table}

\subsubsection{Within-generator rank consistency}\label{sec:results:rankconsist}

\begin{figure}[htbp]
  \centering
  \includegraphics[width=0.9\linewidth]{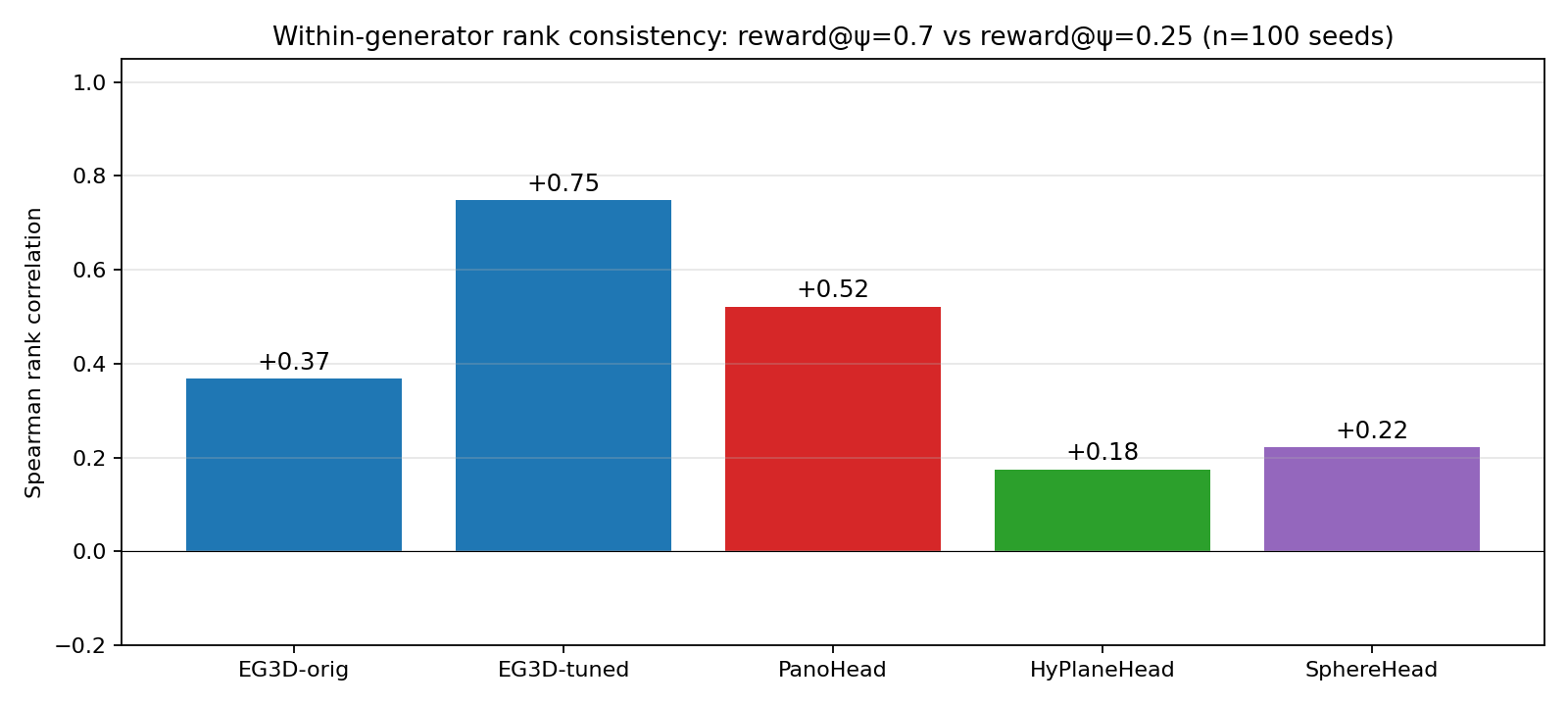}
  \caption{Within-generator rank consistency of the $\sigma_{XYZ}$ reward. For each generator, Spearman rank correlation between per-seed reward at $\psi=0.7$ and at $\psi=0.25$ on the same $100$ latent codes. All values are positive, ranging from $+0.18$ (HyPlaneHead) to $+0.75$ (EG3D-tuned). The strength of within-generator rank stability tracks how much reward-distribution spread the generator retains at low truncation (smaller $\psi=0.7\!\to\!\psi=0.25$ standard-deviation compression $\Rightarrow$ larger rank correlation).}
  \label{fig:within_generator_rank_spearman}
\end{figure}

The more fundamental question is whether the $\sigma_{XYZ}$ reward delivers a stable preference ordering of latent codes inside each generator's distribution. To test, we score the same $100$ latent codes at $\psi=0.7$ and $\psi=0.25$ and compute the Spearman rank correlation between the two per-seed reward sequences for each generator (Figure~\ref{fig:within_generator_rank_spearman}, Table~\ref{tab:within_generator_rank}). The headline result is simple: all five generators remain positively rank-consistent under this diagnostic, but the strength of that consistency varies substantially. EG3D-tuned is the most stable (Spearman $\rho = +0.75$, with $100\%$ of the top-$10$ at $\psi=0.7$ remaining in the top-$50$ at $\psi=0.25$); PanoHead and EG3D-orig are intermediate ($\rho = +0.52$ and $+0.37$); HyPlaneHead and SphereHead are notably weak ($\rho = +0.18$ and $+0.22$).

The pattern tracks the reward-distribution compression from $\psi=0.7$ to $\psi=0.25$. EG3D-orig and EG3D-tuned compress modestly ($1.5\times$ and $1.9\times$ reduction in standard deviation respectively), preserving most of the rank signal. PanoHead compresses more aggressively ($3.1\times$) but still retains usable rank ordering. HyPlaneHead and SphereHead compress by $5\times$ and $4.6\times$ respectively: at $\psi=0.25$ their reward distributions collapse to $\mathrm{std}\!=\!0.08$ and $0.12$ respectively, against canonical-$\psi=0.7$ standard deviations of $0.38$ and $0.55$. When the post-truncation reward spread approaches the reward model's own noise floor, the rank signal is dominated by that noise and rank consistency falls.

The takeaway: the $\sigma_{XYZ}$ reward gives a moderately rank-stable preference ordering on EG3D and PanoHead samples and a weaker but still positive ordering on HyPlaneHead and SphereHead samples. In every case the rank correlation is positive - the reward is not arbitrary within a generator's distribution. But the strength of that ordering depends on how much reward spread the generator's $\sigma$ field retains across truncation, which varies with the generator architecture.

\begin{table}[t]
  \centering
  \caption{Within-generator rank consistency of the $\sigma_{XYZ}$ reward across truncation regimes (the same $100$ latent codes scored at $\psi=0.7$ and at $\psi=0.25$). All five generators retain a positive within-generator rank signal under this diagnostic. Spearman $\rho$ is the rank correlation. The two right-hand columns report how often the top-$10$ / bottom-$10$ latent codes at $\psi=0.7$ remain in the top-$50$ / bottom-$50$ at $\psi=0.25$. ``std-ratio'' is the ratio of the reward standard deviation at $\psi=0.7$ to that at $\psi=0.25$; larger ratios indicate the generator's reward distribution collapses more under truncation and the rank signal degrades.}
  \label{tab:within_generator_rank}
  \footnotesize
  \begin{tabular*}{\tblwidth}{@{}LLLLL@{}}
    \toprule
    Generator & $\rho$ & \makecell[l]{top-$10@0.7$\\$\in$top-$50@0.25$} &
    \makecell[l]{bot-$10@0.7$\\$\in$bot-$50@0.25$} &
    \makecell[l]{std-ratio\\($\psi 0.7 / 0.25$)} \\
    \midrule
    EG3D-orig    & $+0.37$ & $80\%$  & $70\%$  & $1.5\times$ \\
    EG3D-tuned   & $+0.75$ & $100\%$ & $80\%$  & $1.9\times$ \\
    PanoHead     & $+0.52$ & $100\%$ & $100\%$ & $3.1\times$ \\
    HyPlaneHead  & $+0.18$ & $60\%$  & $80\%$  & $5.0\times$ \\
    SphereHead   & $+0.22$ & $60\%$  & $80\%$  & $4.6\times$ \\
    \bottomrule
  \end{tabular*}
\end{table}

\subsubsection{Top-versus-bottom $\sigma_{XYZ}$ mesh tails}\label{sec:results:meshtails}
\begin{figure}[htbp]
  \centering
  \begin{subfigure}{\linewidth}\centering
    \includegraphics[width=0.46\linewidth]{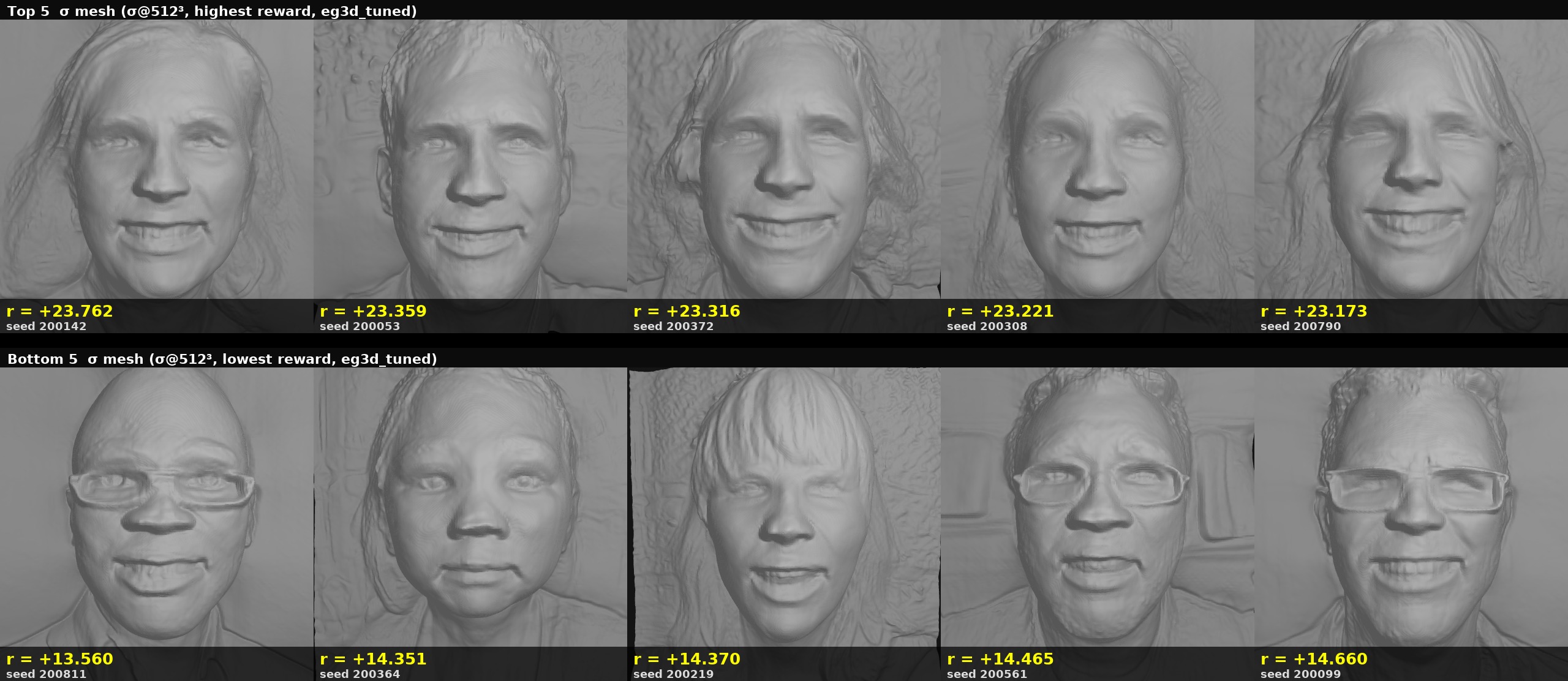}
    \caption{EG3D-tuned}
  \end{subfigure}
  \\[6pt]
  \begin{subfigure}{\linewidth}\centering
    \includegraphics[width=0.46\linewidth]{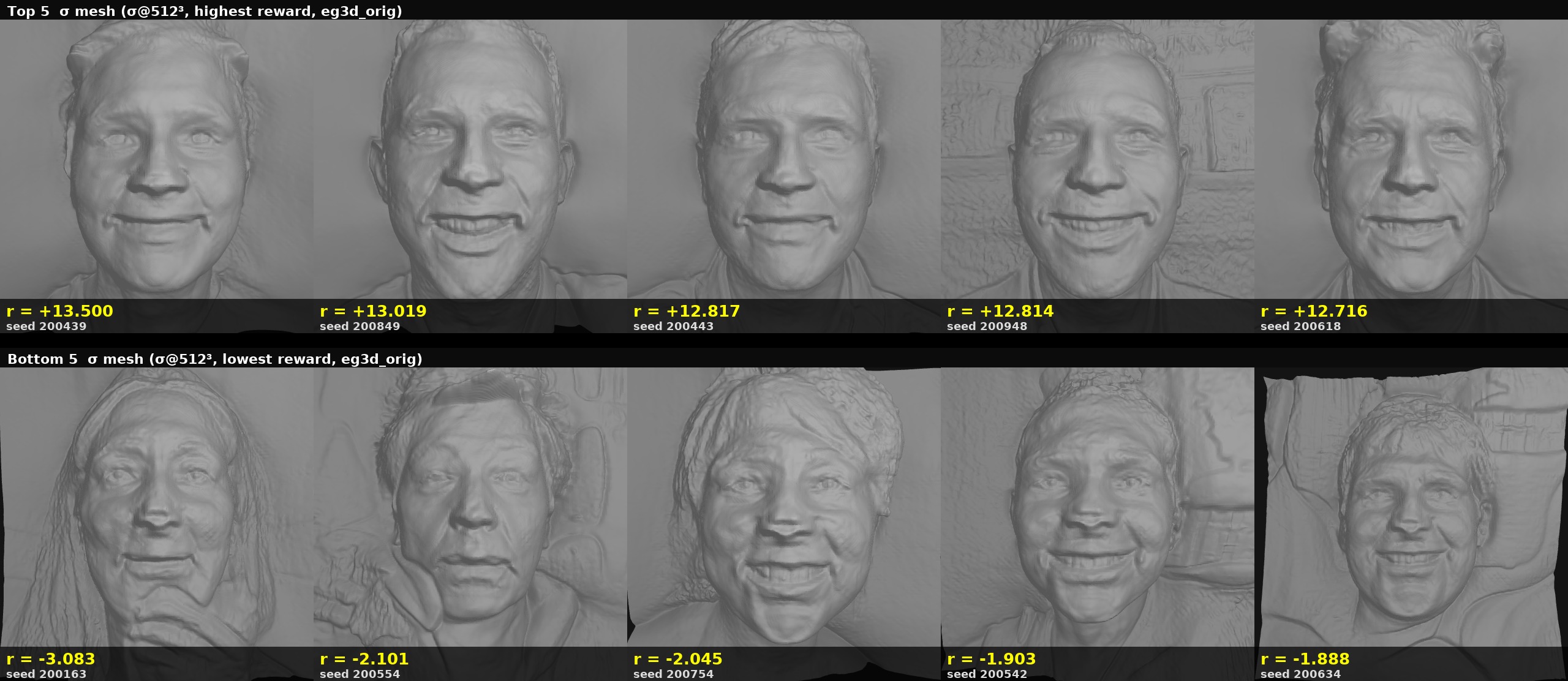}
    \caption{EG3D-orig}
  \end{subfigure}
  \\[6pt]
  \begin{subfigure}{\linewidth}\centering
    \includegraphics[width=0.46\linewidth]{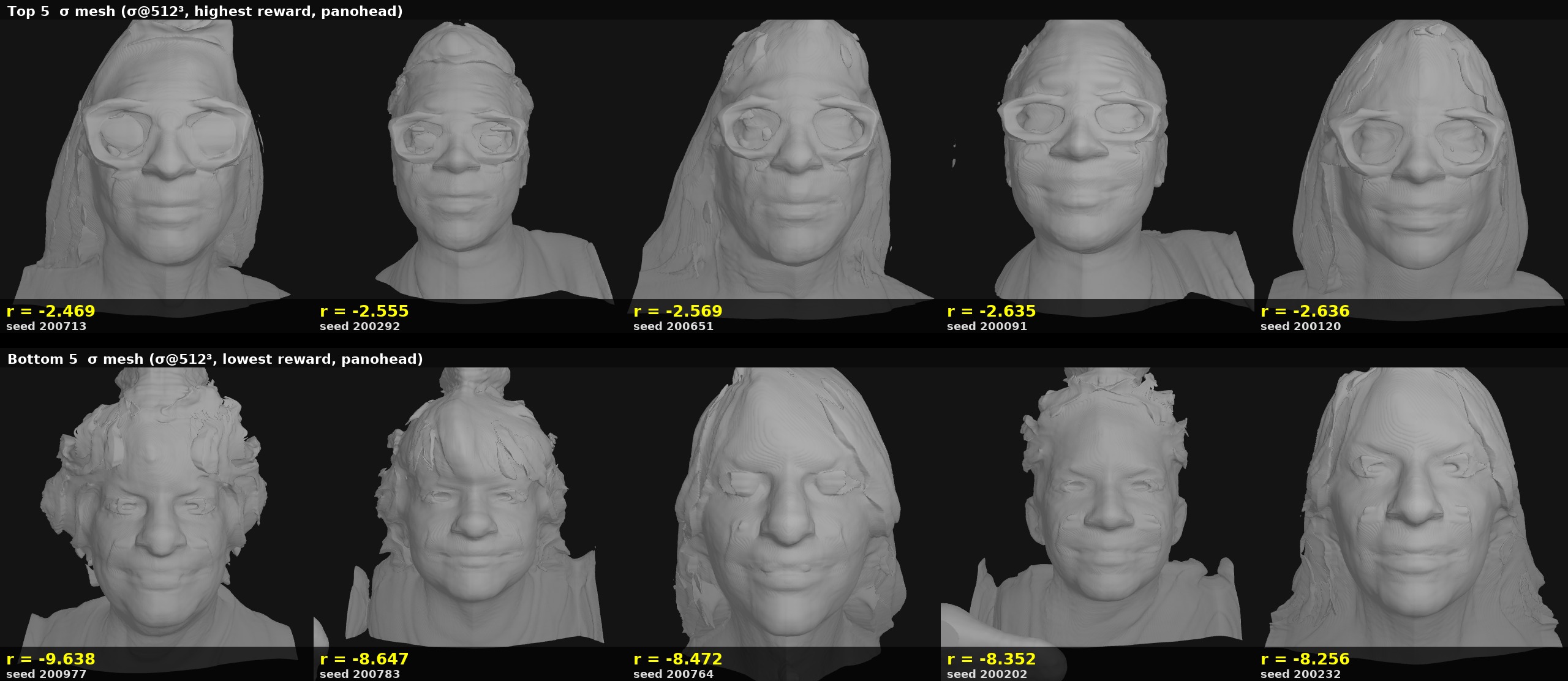}
    \caption{PanoHead}
  \end{subfigure}
  \\[6pt]
  \begin{subfigure}{\linewidth}\centering
    \includegraphics[width=0.46\linewidth]{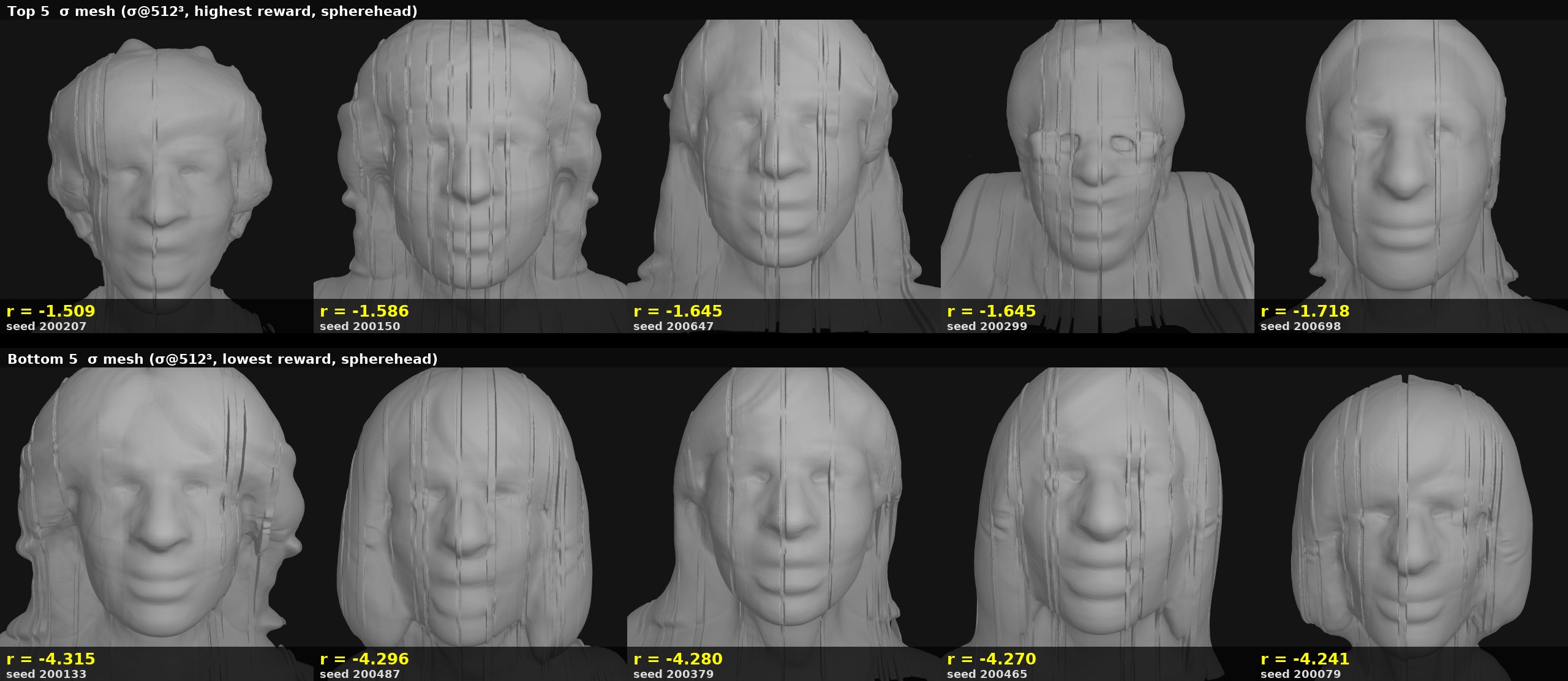}
    \caption{SphereHead}
  \end{subfigure}
  \\[6pt]
  \begin{subfigure}{\linewidth}\centering
    \includegraphics[width=0.46\linewidth]{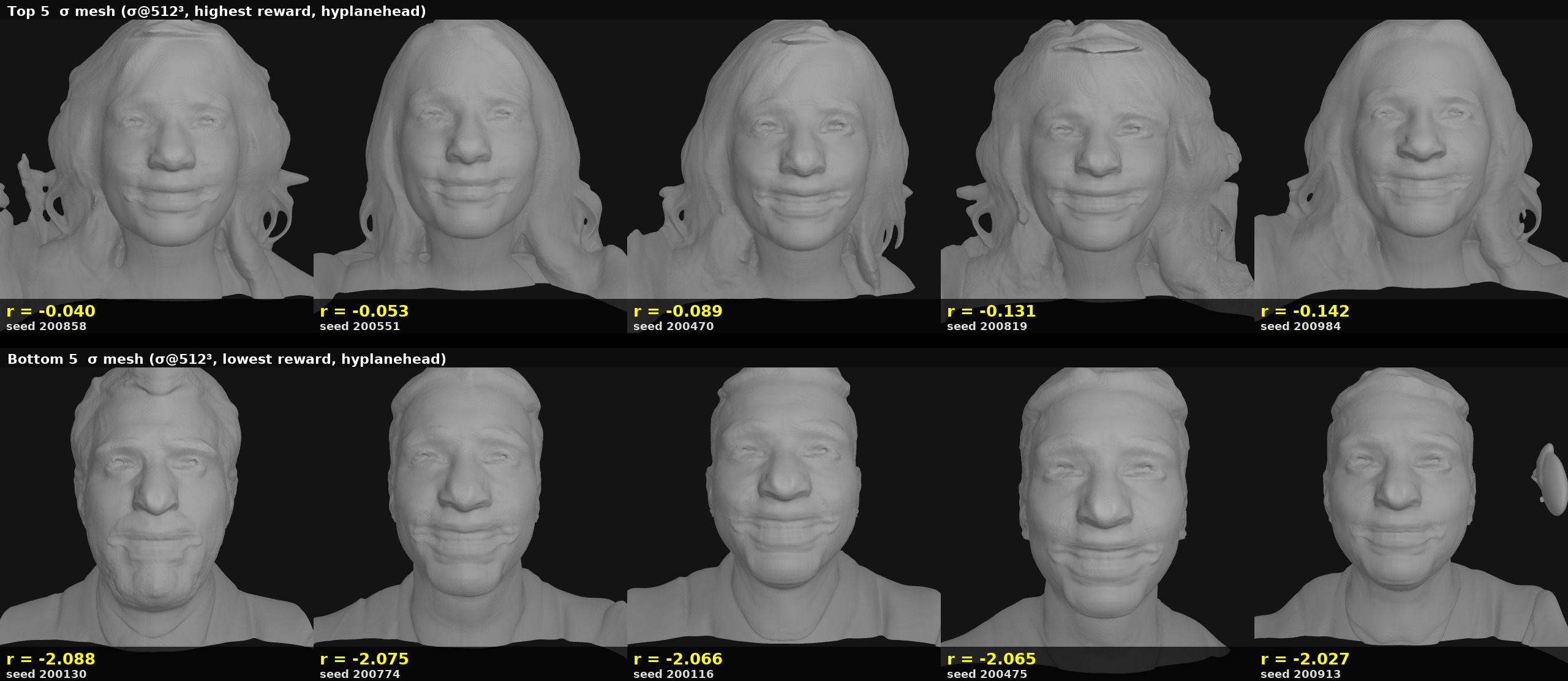}
    \caption{HyPlaneHead}
  \end{subfigure}
  \caption{Unstratified top-$5$ (upper strip) vs bottom-$5$ (lower strip) mesh tails by $\sigma_{XYZ}$ reward, one row of strips per generator (panels (a)--(e)). $\sigma$ is sampled at $512^3$ in memory per seed and marching-cubes-extracted at level $10$. On the two EG3D generators (in-domain for the reward) the top/bottom meshes differ visibly in surface integrity; on the three $360^{\circ}$ generators (out-of-domain for the reward) the top/bottom meshes are not visibly distinguishable beyond demographic correlates. Note that on EG3D-tuned both top-$5$ and bottom-$5$ are of decent quality: the entire tuned distribution sits at or above the EG3D-orig maximum, evidencing that RLHF has lifted the floor of the generator's quality distribution into the in-distribution high-quality regime (bottom-$5$ tuned $\approx$ top-$5$ untuned).}
  \label{fig:unstratified_mesh_tails}
\end{figure}

The within-generator rank-consistency analysis above is purely statistical; it does not directly answer whether the top-ranked samples in each generator's reward distribution look better than the bottom-ranked ones, or whether the reward signal in the cross-architecture generators is meaningful at all. To test this we extend the seed bank to $N{=}1000$ per generator and render the top-$5$ and bottom-$5$ samples by $\sigma_{XYZ}$ reward as marching-cubes meshes at $\sigma$-resolution $512^3$ (in-memory only; Figure~\ref{fig:unstratified_mesh_tails}). The reward dynamic ranges between top-$5$ and bottom-$5$ are EG3D-tuned $\Delta r \approx 10.2$ ($+23.76 \to +13.56$), EG3D-orig   $\Delta r \approx 16.6$ ($+13.50 \to -3.08$), PanoHead    $\Delta r \approx 7.2$ ($-2.47 \to -9.64$), SphereHead  $\Delta r \approx 2.8$ ($-1.51 \to -4.32$), and HyPlaneHead $\Delta r \approx 2.0$ ($-0.04 \to -2.09$).

The qualitative gap between top and bottom mesh tails differs sharply across the two EG3D generators in a way that is itself informative (Figure~\ref{fig:unstratified_mesh_tails}, EG3D rows). On EG3D-orig the gap is large: high-reward meshes show clean canonical-frontal face surfaces with articulated eye, nose and mouth geometry, while low-reward meshes are visibly degenerate (collapsed surfaces, asymmetric density, broken forehead/cheek topology). On EG3D-tuned, by contrast, the gap is small because the bottom-$5$ meshes are themselves of decent geometric quality - both ends of the tuned-model reward distribution sit in the high-quality regime. The numerical evidence for this is that the worst tuned sample's reward ($+13.56$) is approximately the best untuned sample's reward ($+13.50$): the entire EG3D-tuned distribution sits at or above the EG3D-orig maximum. 3D reward tuning has not merely shifted the mean upward, it has lifted the floor of the generator's quality distribution into the regime that previously existed only in the right-tail of EG3D-orig. The bottom-$5$ of the tuned model is comparable to the top-$5$ of the untuned model.

\begin{figure}[htbp]
  \centering
  \includegraphics[width=\linewidth]{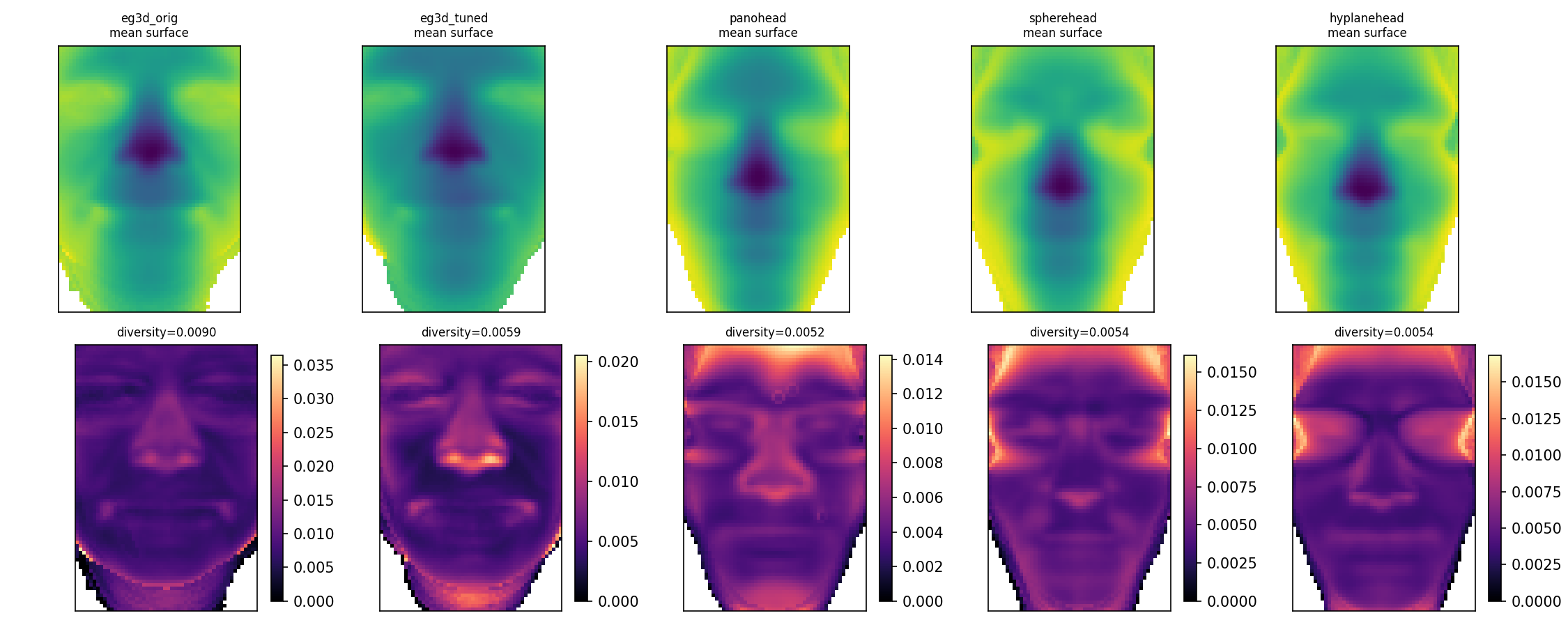}
  \caption{Canonical-view facial depth diagnostics across the five generators. Top row: mean ray-termination depth within the fixed facial window. Bottom row: across-seed depth variation in the same window, with the scalar diversity score reported above each panel. EG3D-orig shows the strongest interior facial relief and the largest concentration of variation around semantically meaningful facial features, especially the eyes, nose and mouth. EG3D-tuned remains structured but is visibly smoother. PanoHead, SphereHead and HyPlaneHead cluster at lower diversity and exhibit flatter facial interiors, helping explain why the reward has a weaker within-domain ordering signal on those architectures under the present diagnostic.}
  \label{fig:facial_depth_diversity_diag}
\end{figure}

On the three $360^{\circ}$ generators the top and bottom mesh tails are visually far more similar to one another than on EG3D. We quantify this directly, and Figure~\ref{fig:facial_depth_diversity_diag} makes the effect visually explicit: EG3D-orig shows stronger within-face relief and more pronounced variation concentrated around the eyes, nose, mouth and cheek contours, whereas PanoHead, SphereHead and HyPlaneHead are flatter through the facial interior and exhibit less differentiated variation away from the boundary bands. Measuring facial-surface diversity as the across-seed variation of the canonical-view ray-termination depth within a fixed facial window - a surface measure that is comparable across architectures because it reads the depth at which each ray terminates rather than thresholding the raw $\sigma$ field, whose scale differs markedly between these NeRFs - EG3D-orig is by far the most diverse ($0.0090$), fine-tuned EG3D is intermediate ($0.0059$), and PanoHead, SphereHead and HyPlaneHead cluster low and close together ($0.0052$, $0.0054$, $0.0054$). Under the current reward and crop convention, the compressed reward range on the $360^{\circ}$ generators ($\Delta r \le 7.2$ vs $\Delta r \approx 16.6$ for EG3D-orig) is consistent with those generators exposing less facial geometric variation to the reward than EG3D does. The cross-architecture rank signal documented in Table~\ref{tab:within_generator_rank} (Spearman $\rho \in [0.18, 0.52]$ for the $360^{\circ}$ generators) is real but correspondingly weak. As an honest caveat, two pretrained image-based reward models (MVReward and Reward3D) rank these same geometries differently from $\sigma_{XYZ}$, with within-generator rank correlations near zero. This can be read not as a contradiction but as further evidence that density-field geometric quality is not captured by these 2D multi-view image rewards.

The practical implication is twofold. Fine-tuning EG3D with the $\sigma_{XYZ}$ reward works because the reward orders genuine geometric quality within EG3D's $\sigma$ distribution (Figure~\ref{fig:unstratified_mesh_tails}, EG3D rows). In addition, porting the same reward to a new generator architecture may require retraining on that architecture's $\sigma$ distribution in order to recover a stronger within-domain quality ranking. Cross-architecture reward transfer may be bounded by the alignment of the target generator's $\sigma$ representation with the reward's training distribution. 

\subsubsection{Reward-guided inversion on SphereHead}\label{sec:results:spherehead_inversion}
\begin{figure}[htbp]
  \centering
  \includegraphics[width=\linewidth]{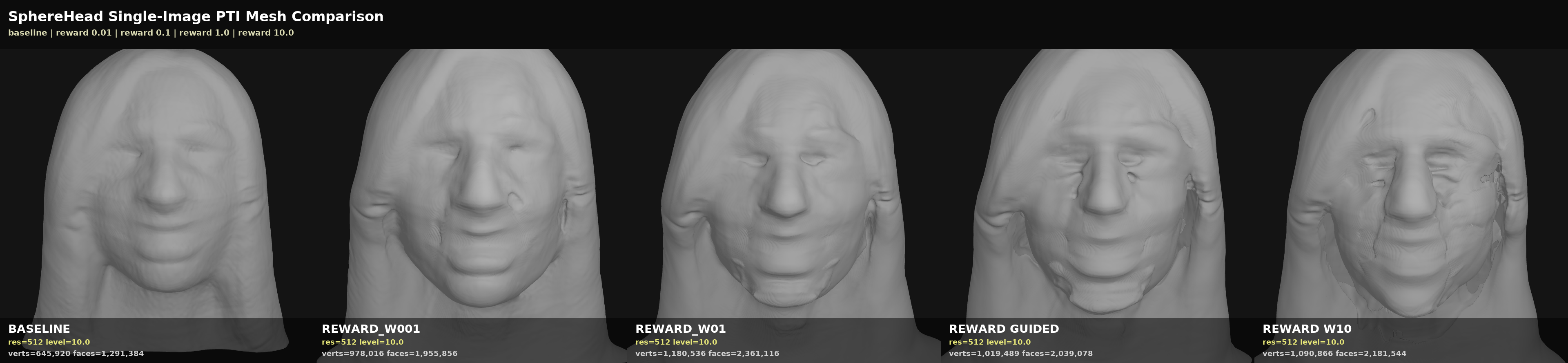}
  \caption{Single-image PTI inversion of SphereHead with EG3D-reward guidance at increasing weight (left to right: baseline, then $w=0.01, 0.1, 1, 10$), shown as marching-cubes meshes ($512^3$, level $10$). The SphereHead baseline is smoother and less detailed than EG3D, and adding the EG3D-trained reward distorts the surface (blistered, irregular geometry) in every case rather than refining it, confirming that the reward does not transfer cleanly to SphereHead under this setup.}
  \label{fig:spherehead_inversion}
\end{figure}

As a direct test of whether the $\sigma_{XYZ}$ reward can improve a different generator's geometry, we run single-image pivotal-tuning inversion (PTI) \citep{roich2022pti} on SphereHead while adding the EG3D-trained reward as a guidance term, sweeping its weight $w \in \{0, 0.01, 0.1, 1, 10\}$ (Figure~\ref{fig:spherehead_inversion}, Table~\ref{tab:spherehead_inversion}). Increasing $w$ raises the reward score monotonically (from $6.69$ at $w=0.01$ to $8.56$ at $w=10$) but steadily worsens the image reconstruction (MSE $0.032 \to 0.069$, perceptual $0.099 \to 0.186$, both above the $0.027 / 0.091$ baseline). Crucially, the extracted meshes show that the reward does not refine the SphereHead surface so much as distort it: the SphereHead baseline is already smoother and less detailed than EG3D, and every reward-guided run develops blistered, irregular geometry, with no weight recovering the selective, appearance-preserving improvement seen on EG3D. Under this inversion setup, the result is consistent with the rest of this section: the reward is bound to EG3D's $\sigma$ distribution and does not transfer cleanly to SphereHead, whose $\sigma$ field is out-of-distribution (Figure~\ref{fig:sigma_histogram}) and whose facial geometry shows relatively low diversity under the present depth-based diagnostic.

\begin{table}[t]
  \centering
  \caption{Single-image PTI inversion on SphereHead with EG3D reward guidance at weight $w$. A higher weight raises the reward score but worsens image fit (lower MSE / perceptual is better); the geometry is distorted rather than refined (Figure~\ref{fig:spherehead_inversion}).}
  \label{tab:spherehead_inversion}
  \footnotesize
  \begin{tabular*}{\tblwidth}{@{}LLLL@{}}
    \toprule
    Reward weight $w$ & MSE & Perceptual & Reward \\
    \midrule
    $0$ (baseline) & 0.027 & 0.091 & - \\
    $0.01$         & 0.032 & 0.099 & 6.69 \\
    $0.1$          & 0.038 & 0.124 & 7.32 \\
    $1.0$          & 0.055 & 0.162 & 7.23 \\
    $10.0$         & 0.069 & 0.186 & 8.56 \\
    \bottomrule
  \end{tabular*}
\end{table}

\section{Discussion}\label{sec:discussion}

The fine-tuning results suggest that a reward model using the $\sigma_{XYZ}$ features is particularly sensitive to geometric issues in desired regions such as the nose, face sides, and forehead. This is learned from simple rankings in a weakly supervised manner, without requiring direct annotation of problematic geometries. Compared with the contemporary wave of preference-driven 3D generative methods reviewed in Section~\ref{sec:related}, the present setting differs in three connected ways already established in Section~\ref{sec:rel:position}. Our reward model evaluates a continuous density field rather than rendered multi-view imagery \citep{ye2024dreamreward,zhou2025dreamdpo,wang2025mvreward} or mesh tokens \citep{zou2026dreamcs,zhao2025deepmesh,liu2025meshrft}, avoiding view-dependent reward bias and the discretisation step inherent to mesh extraction. The absence of a text prompt decouples the reward signal from the joint $(\text{prompt}, 3\text{D})$ embedding that conditions contemporary preference-tuned methods \citep{zou2026dreamcs}; fine-tuning with our reward improves the geometry while a density-consistency loss $\mathcal{L}_{c}$ keeps the 2D appearance qualitatively similar at bounded cost in FID shift. And the preference set is small, approximately an order of magnitude smaller than DeepMesh \citep{zhao2025deepmesh} and over four times smaller than MVReward \citep{wang2025mvreward}, from a single annotator, which is feasible precisely because the reward operates in the unconditional setting and need not disentangle text-conditioned semantics.

The post-hoc analyses of Section~\ref{sec:results:posthoc} bear on several aspects of whether the fine-tuned representation is merely simplified across image and 3D features. The mode-collapse concern arising from the loss of FID after $20$ kimg of fine-tuning is partially addressed by the truncation-baseline comparison of Section~\ref{sec:results:truncation}: tuned samples are closer to the original than to the $\psi = 0$ mean face in $93\%$--$98\%$ of cases, and most of the geometric change is orthogonal to the truncation axis - ruling out the worst form of collapse, in which every sample is pulled toward a common mean shape. The open question of why the $\sigma_{XYZ}$ representation outperforms depth maps and point clouds is partially answered by the region-attribution analysis of Section~\ref{sec:results:shapley_face}: on $\sigma_{XYZ}$ the reward model attends predominantly to nose, mouth, and cheeks, whereas on depth maps and point clouds it attends to face edges away from these features. The identity-drift concern is addressed by the matched-identity reward analysis of Section~\ref{sec:results:robustness} and the identity-vs-geometry decorrelation of Section~\ref{sec:results:truncation}: reward improvement persists when identity is held fixed externally, and the small residual identity drift (mean canonical cosine $0.84$, worst-view $0.72$; view-consistency drops of $0.010$--$0.017$) is statistically resolvable but small and approximately uncorrelated with the magnitude of the geometric change.

Several limitations remain unresolved. The most salient is that the reward is bound to the generator on which it was trained: it does not retain its dynamic range when transferred to the PanoHead, SphereHead and HyPlaneHead families. The current diagnostics suggest two main contributing factors. The $\sigma$ fields of those generators occupy a very different numerical regime from EG3D-FFHQ's - their peak densities differ by more than an order of magnitude (Figure~\ref{fig:sigma_histogram}) - so an EG3D-trained reward is evaluated out-of-distribution. In addition, their facial geometries appear less diverse than EG3D's under the present canonical-view depth measure, leaving less geometric variation for any reward to resolve. Recovering the full dynamic range on a new generator would likely require re-training the reward on that generator's own $\sigma$ distribution.

Another limitation concerns supervision. All preference data come from a single annotator, after an initial multi-respondent triplet-ranking attempt failed to converge (Section~\ref{sec:method:dataset}); the study is therefore best read as a proof of concept for density-field preference learning rather than a model of broad inter-annotator human preference. We expect the pipeline to extend to multiple annotators - the reward is cheap to train - but validating agreement across annotators, and the robustness of fine-tuning to a multi-annotator reward, is left to future work. 

Several extensions remain worthwhile. The per-seed comparison of the $\sigma_{XYZ}$ reward against pretrained rendered-image rewards \citep{ye2024dreamreward,wang2025mvreward} reported in Section~\ref{sec:results:image_reward} found that the two image-based rewards correlate with one another but not with $\sigma_{XYZ}$: MVReward is out-of-distribution on FFHQ-domain face renders and uncorrelated with $\sigma_{XYZ}$ ($\rho=-0.05$ on the before/after deltas), while Reward3D is weakly positively aligned ($\rho=+0.25$) and, under the fixed prompt used in that diagnostic, agrees that the tuned generator is improved in $77/100$ seeds. A full GAN-loop replacement of $\mathcal{L}_{r}$ by either image reward remains a worthwhile extension. Mesh-feature rewards \citep{zou2026dreamcs,zhao2025deepmesh} would further isolate the contribution of the reward-input representation from that of the fine-tuning loop, but require either a differentiable iso-surface extractor \citep{neupane2026nerfmesh} or a score-function gradient estimator, which we leave for future work. We have evaluated the reward's diagnostic transfer to the recent face-domain 3D generators PanoHead \citep{an2023panohead}, SphereHead \citep{li2024spherehead} and HyPlaneHead \citep{li2026hyplanehead} (Section~\ref{sec:results:generalise}); extending the fine-tuning itself to those generators, or to image-to-3D pipelines such as Trellis~2 \citep{xiang2025trellis2} and CraftsMan3D \citep{li2025craftsman3d}, would clarify how broadly the approach transfers across 3D representations.

\section{Conclusion}\label{sec:conclusion}

This work demonstrates the potential of using human preferences to fine-tune 3D shapes in an unconditional 3D-aware GAN in a data- and compute-efficient setting. Our approach simultaneously produces a model $r_{\theta}$ which scores quality directly from the radiance-field density values, sidestepping the need for either text-prompt conditioning or mesh extraction. In this single-annotator proof-of-concept setting, the procedure improves 3D quality as judged by external user preference using relatively few training samples, while keeping 2D appearance qualitatively similar at a measurable but bounded cost. Our findings support the notion that human preferences can be used in a fine-tuning stage to improve desired characteristics in an implicit 3D representation alone, decoupling the reliance on text conditioning to learn a preference model from weaker supervision.

\bibliographystyle{plainnat}
\bibliography{references}

\end{document}